\PassOptionsToPackage{table,xcdraw,usenames,dvipsnames}{xcolor}

\documentclass{article}

\usepackage{microtype}
\usepackage{graphicx}
\usepackage{subfigure}
\usepackage{booktabs} % for professional tables

\usepackage{hyperref}

\usepackage[accepted]{icml2025}

\usepackage{amsmath}
\usepackage{amssymb}
\usepackage{mathtools}
\usepackage{amsthm}
\usepackage{xcolor}         % colors
\usepackage{colortbl}
\usepackage{xspace}
\usepackage{arydshln}
\usepackage{relsize}
\usepackage{array}
\usepackage{pifont}
\usepackage{hhline}
\usepackage{boldline}
\usepackage{float}
\usepackage{bbding}
\usepackage{amsfonts}       % blackboard math symbols
\usepackage{nicefrac}       % compact symbols for 1/2, etc.
\usepackage[figuresright]{rotating}
\usepackage{balance}
\usepackage[dvipsnames]{xcolor} 
\usepackage{hyperref}
\theoremstyle{plain}

\theoremstyle{definition}

\theoremstyle{remark}

\usepackage[textsize=tiny]{todonotes}

\usepackage{paralist}
\setdefaultleftmargin{1.2em}{}{}{}{}{}

\newcommand{\VistaDPOtitle}{$\mathcal{V}ista\mathcal{DPO}$ \xspace}

\definecolor{lightgray}{RGB}{242,242,242}

\icmltitlerunning{Video Hierarchical Spatial-Temporal Preference Optimization}

\begin{document}

\twocolumn[
% \icmltitle{\textit{ViSTA}: Mitigating Intrinsic Hallucinations in Video LMMs through\\Hierarchical Spatial-Temporal Preference Optimization}

% \icmltitle{\ViSTAtitle: Mitigating Intrinsic Hallucinations in Video LMMs via\\Hierarchical Spatial-Temporal Preference Optimization}

% \icmltitle{\VistaDPOtitle: Video Hierarchical Spatial-Temporal Direct Preference Optimization for Large Video Models}

\icmltitle{\VistaDPOtitle: Video Hierarchical Spatial-Temporal Direct\\ Preference Optimization for Large Video Models}

% Author list: 

% Haojian Huang* (HKU), 
% Haodong Chen* (HKUST), (*Equal Contribution)
% Shengqiong Wu (NUS)
% Meng Luo (NUS)
% Jinlan Fu (NUS)
% Yuhui Zhang (Stanford)
% Xinya Du (UTD)
% Hao Fei (NUS)  (# Corespondance)

% Hanwang Zhang (NTU)
% Mohit Bansal (UNC Chapel Hill)



\icmlsetsymbol{equal}{*}

\begin{icmlauthorlist}
\icmlauthor{Haojian Huang}{equal,hku}
\icmlauthor{Haodong Chen}{equal,hkust}
\icmlauthor{Shengqiong Wu}{nus}
\icmlauthor{Meng Luo}{nus}\\
\icmlauthor{Jinlan Fu}{nus}
\icmlauthor{Xinya Du}{utd}
\icmlauthor{Hanwang Zhang}{ntu}
\icmlauthor{Hao Fei}{nus}
\end{icmlauthorlist}

\icmlaffiliation{hku}{The University of Hong Kong}
\icmlaffiliation{hkust}{The Hong Kong University of Science and Technology}
\icmlaffiliation{nus}{National University of Singapore}
\icmlaffiliation{utd}{University of Texas at Dallas}
\icmlaffiliation{ntu}{Nanyang Technological University}

\icmlcorrespondingauthor{Hao Fei}{haofei37@nus.edu.sg}

\icmlkeywords{Machine Learning, ICML}

\vskip 0.3in
]



\printAffiliationsAndNotice{\icmlEqualContribution} % otherwise use the standard text.

\begin{abstract}
Large Video Models (LVMs) built upon Large Language Models (LLMs) have shown promise in video understanding but often suffer from misalignment with human intuition and video hallucination issues. 
To address these challenges, we introduce \textbf{VistaDPO}, a novel framework for Video Hierarchical Spatial-Temporal Direct Preference Optimization. VistaDPO enhances text-video preference alignment across three hierarchical levels: 
i) \textbf{\em Instance Level}, aligning overall video content with responses; 
ii) \textbf{\em Temporal Level}, aligning video temporal semantics with event descriptions; 
and iii) \textbf{\em Perceptive Level}, aligning spatial objects with language tokens. 
Given the lack of datasets for fine-grained video-language preference alignment, we construct \textbf{VistaDPO-7k}, a dataset of 7.2K QA pairs annotated with chosen and rejected responses, along with spatial-temporal grounding information such as timestamps, keyframes, and bounding boxes.  
Extensive experiments on benchmarks such as Video Hallucination, Video QA, and Captioning performance tasks demonstrate that VistaDPO significantly improves the performance of existing LVMs, effectively mitigating video-language misalignment and hallucination.
% We open our code and data (\url{https://anonymous.4open.science/r/VistaDPO}) later.
The code and data are available at \textcolor{blue}{\href{https://github.com/HaroldChen19/VistaDPO}{VistaDPO Repository}}.
\end{abstract}

\vspace{-8mm}

\section{Introduction}
\label{sec:intro}

\vspace{-1mm}

\begin{figure}[!t]
  \centering
  \vspace{-1em}
   \includegraphics[width=0.99\linewidth]{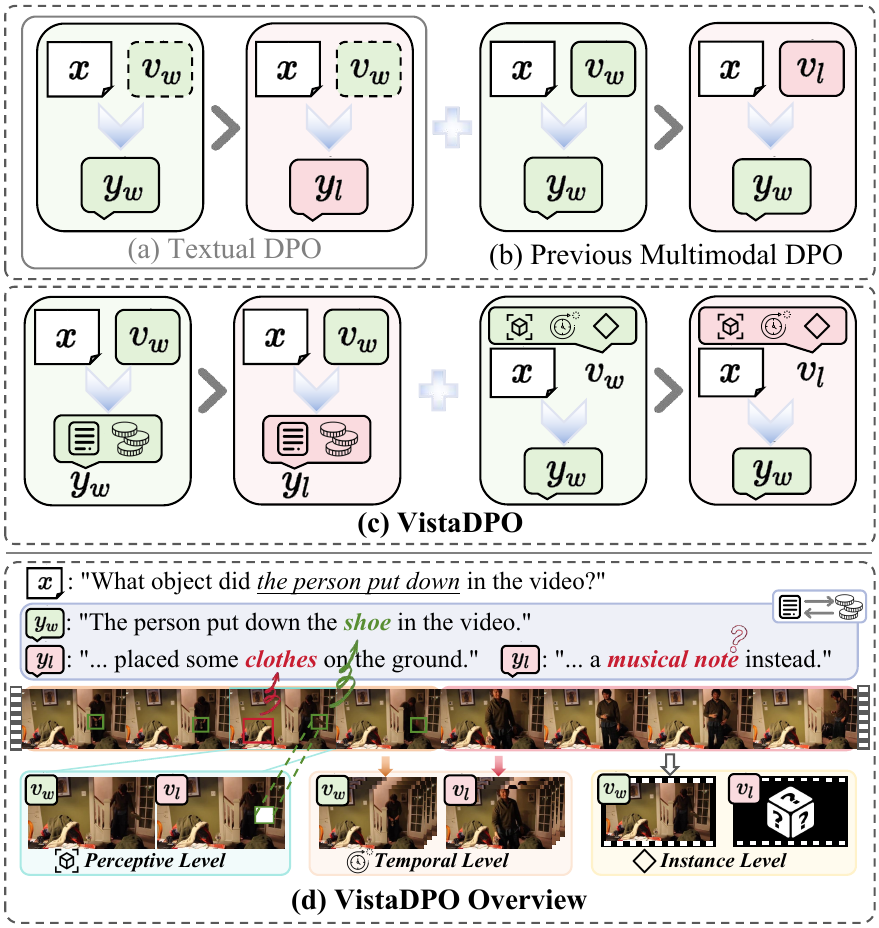}
   \vspace{-1em}
   \caption{(a) Traditional textual DPO overlooks multimodal information, limiting video-language tasks. (b) Existing multimodal DPO methods rely on coarse alignment, missing rich temporal and perceptual details. (c\&d) \textbf{VistaDPO} overcomes these limitations with a hierarchical spatiotemporal preference optimization framework, enabling fine-grained video-language alignment and precise reasoning over video dynamics. Here, $y_w$ is the preferred response over $y_l$, and $v_w$ the visual input more likely to produce it than $v_l$.}
   \vspace{-5mm}
   \label{fig:intro}
\end{figure}

Achieving human-like reasoning capabilities for videos is a critical research topic in the field of AI.
In recent years, Large Video Models (LVMs) \cite{li2023videochat,zhang2023video,lin2023video, li2024mvbench,wu24next,cheng2024videollama,fei2024vitron,jin2024chat,qian2024momentor,li2025llama} have garnered significant research attention. 
Built upon Large Language Models (LLMs) \cite{touvron2023llama,bai2023qwen,peng2023instruction,dubey2024llama}, LVMs leverage the powerful intelligence of LLMs in language, achieving unprecedented understanding of video content.
However, increasing studies reveal that LVMs encounter critical issues, such as video understanding that deviates from human intuition \cite{zhou2024humanvbench,fei2024video,cheng2024videgothink,hu2024recent} or the phenomenon of video hallucination \cite{wang2024videohallucer,yuan2024videorefer, ma2024beyond}, where the model outputs content that does not align with the input, \textit{e.g.}, user instructions, video content.
% , or factual information.
The root of these issues lies in the inherent nature of current LVM architectures \cite{yan2021videogpt,cheng2024videollama,lin2023video}, where most LVMs integrate a video encoder (\textit{e.g.}, ViT) into text-oriented LLMs through a connector to achieve video signal interpretation.
Since backbone LLMs undergo extensive pre-training on large-scale language data while video encoders lack peer capability, this gap leads LLMs to produce overly confident outputs based on biased or even incorrect perceptions of video content from the encoder.
While the supervised fine-tuning (SFT) with video-language pairs \cite{wang2024videohallucer,leng2024curse,yuan2024videorefer} can partially improve the alignment between the two modalities in LVMs, fundamentally addressing the issue requires reliance on extremely large-scale data.

Recently, Direct Preference Optimization (DPO) \cite{rafailov2024direct} has been proposed as a promising alternative to SFT. 
It trains LLMs to prefer responses chosen by evaluators over rejected ones when presented with a user query.
By identifying which response better aligns with human preferences rather than requiring precise target outputs, DPO significantly alleviates dependence on annotated data while enhancing alignment with human values and effectively addressing hallucination issues. 
% DPO requires only identifying which response better aligns with human preferences, without the need for precise target outputs. This significantly alleviates the dependency on annotated data while greatly enhancing the alignment of models with human values, effectively addressing hallucination issues.
Some follow-up studies \cite{xie2024v,liu2024mia,zhou2024aligning,fu2025chip} have extended DPO from textual to multimodal LLMs, facilitating cross-modal alignment and improving the generalization capabilities of the models.
Most recently, Hound-DPO \cite{zhang2024direct} pioneers a video DPO, demonstrating that tailored rewards through DPO can significantly enhance the performance of LVMs.
Unfortunately, we find that this work straightforwardly applies the DPO strategy designed for image-text LLMs to video-language preference alignment (as shown in Figure~\ref{fig:intro}), which introduces two critical limitations.
% their straightforward adaptation of text/image DPO strategies to video-language preference alignment, as shown in Figure~\ref{fig:1}, introduces two significant limitations. 
\textbf{First}, \citet{zhang2024direct} fails to adequately consider the temporal characteristics of videos.
Unlike static images, videos always require both spatial semantic understanding and dynamic temporal reasoning \cite{fei2024enhancing}, necessitating a comprehensive modeling of the spatial-temporal attributes of videos.
\textbf{Second}, their work focuses solely on coarse-grained alignment between video and language (response text) at the instance level, which may lead to suboptimal preference alignment \cite{zeng2024token, gunjal2024detecting}.
We emphasize that achieving proper alignment between two modalities requires a fine-grained preference alignment. 
Intuitively, dynamic videos correspond to paired text at multiple hierarchical levels.

To address these challenges, we propose a novel framework, \emph{Video Hierarchical Spatial-Temporal Direct Preference Optimization} (namely \textbf{\texttt{VistaDPO}}), aiming to strengthen LVMs.
VistaDPO improves text-video preference alignment across hierarchical granularities.
Specifically, we design three levels of alignment (as shown in Figure~\ref{fig:intro}): 
\vspace{-2mm}
\begin{compactitem}
    \item[$\blacktriangleright$] \textbf{Instance Level:} Matching the overall video content with the most appropriate response for semantic alignment.
    \item[$\blacktriangleright$] \textbf{Temporal Level:} Aligning video temporal semantics with event descriptions, enabling temporal reasoning. 
    \item[$\blacktriangleright$] \textbf{Perceptive Level:} Aligning video spatial objects (\textit{i.e.}, regions of interest) with objective tokens or phrases in the language at a fine-grained semantic level.  
\end{compactitem}
\vspace{-2mm}
To implement such fine-grained preference optimization, we construct a large-scale spatial-temporally grounded video dataset called \textbf{\texttt{VistaDPO-7k}}.
We manually annotate 3,878 videos with spatial-temporal groundings in a video QA format, providing high-quality labels for hallucinated and non-hallucinated answers, along with timestamps, keyframes, and bounding boxes of relevant semantics.

We conduct extensive evaluation on benchmarks including Video Hallucination, Video QA, Captioning Tasks, by post-training existing popular LVMs with the proposed VistaDPO.
The results show that VistaDPO consistently improves baseline LVMs, achieving significant average improvements of $26.42$\% over PLLaVA and $53.92$\% over Video-LLaVA respectively.
Through in-depth analysis, we show that VistaDPO effectively and comprehensively captures the dynamic interactions between video content and texts, thanks to its hierarchical spatial-temporal alignment strategy. To summarize, this work contributes in threefold:
\vspace{-2mm}
\begin{compactitem}
    \item Propose a novel Video Hierarchical Spatial-Temporal DPO (\textbf{VistaDPO}) mechanism, a more fine-grained DPO strategy to optimize the alignment between video and language in LVMs.  
    \item Construct and release a large-scale (7.2K) high-quality annotated QA pairs dataset, which can serve as a valuable resource for follow-up video DPO research.  
    \item Empirically, VistaDPO significantly improves the generalization capabilities of existing LVMs, effectively mitigating video-language misalignment and hallucination.
\end{compactitem}
\vspace{-0.8em}

\section{Related Work}
\label{sec:related_work}

% \textbf{Hallucination in LMMs}

% \lipsum[1]
% [1-10]

% \textbf{DPO for LLMs}

% \lipsum[2]

% \lipsum[3][1-18]

% \subsection{LLMs/MLLMs and Their Issues}

By building on powerful LLMs and integrating various multimodal encoders, researchers have developed MLLMs \cite{liu2024visual, fu2025vita, yin2024woodpecker, wu2024deepseek} and LVMs \cite{li2023videochat,zhang2023video,lin2023video, li2024mvbench,cheng2024videollama,jin2024chat,li2025llama}.
Through necessary SFT on visual instruction-tuning data, MLLMs and LVMs have not only developed robust multimodal understanding capabilities but have also significantly enhanced human-computer interaction, making cross-modal interactions more intuitive and seamless.
Unfortunately, inheriting the intrinsic hallucination issues of LLMs, LVMs also frequently suffer from hallucinations \cite{liu2024survey, zhang2024direct, li2024vidhalluc, liu2024seeing} or fail to align their understanding of visual content with human values.
Increasing the volume of multimodal SFT data has been shown to alleviate these issues to some extent \cite{ahn2024tuning, tan2024evalalign, jiang2024supervised,chen2024finecliper}. 
However, this approach is often accompanied by higher annotation costs and computational expenses.
This challenge is particularly pronounced in video scenarios, where LVMs demand significantly larger datasets and higher training costs.

% \subsection{From DPO to Video DPO}

Subsequently, the community has introduced the DPO technique \cite{rafailov2024direct}, where preference alignment aligns LLMs with human values, reducing hallucinations by guiding the model's adjustments using pairs of preferred and rejected data.
Multimodal preference alignment, as an extension of preference alignment techniques to visual and textual inputs, has been widely applied to MLLMs to improve cross-modal alignment \cite{liu2024mia, xie2024v, zhou2024aligning} as shown in Table~\ref{app_tab:dpo_comparison}.
Recently, Hound-DPO, pioneered by \citet{zhang2024direct}, successfully applies multimodal DPO to LVMs, improving video understanding and addressing hallucination issues. However, it overlooks the preference alignment of visual inputs.
In this paper, we aim to further enhance the effectiveness of DPO in video scenarios by modeling fine-grained alignments between video and language. To achieve this, we propose a hierarchical preference optimization framework that efficiently captures dynamic spatial-temporal dependencies in video tasks.

\vspace{-3mm}
\section{Preliminaries}
\label{sec:preliminaries}

\vspace{-1mm}
Direct Preference Optimization (DPO) \cite{rafailov2024direct} aligns language models with human preferences, removing the need for explicit reward modeling or reinforcement learning (RL). Given a model $\pi_\theta$ (the target model) and a reference policy $\pi_{\mathrm{ref}}$ (from supervised fine-tuning), the RL objective in reinforcement learning with human feedback (RLHF), initialized with $\pi_\theta=\pi_{\mathrm{ref}}$, is expressed as:
\setlength\abovedisplayskip{3pt}
\setlength\belowdisplayskip{3pt}
\begin{equation}
\begin{aligned}
\max_{\pi_\theta}\mathbb{E}&_{x\thicksim\mathcal{D},y\thicksim\pi_\theta(y|x)}\begin{bmatrix}r(x,y)\end{bmatrix}\\&-\beta\mathbb{D}_{\mathrm{KL}}\begin{bmatrix}\pi_\theta(y\mid x)\parallel\pi_{\mathrm{ref}}(y\mid x)\end{bmatrix}, \label{eq1}
\end{aligned}
\end{equation}
where $r(x,y)$ denotes the reward function with $x$ as the input instruction and $y$ as the response. DPO establishes a mapping between the reward model and the optimal policy under the reverse KL divergence, obtaining a representation of the reward function concerning the policy:
\setlength\abovedisplayskip{3pt}
\setlength\belowdisplayskip{3pt}
\begin{equation}
    r(x,y)=\beta\log\frac{\pi_\theta(y|x)}{\pi_{\mathrm{ref}}(y|x)}+\beta\log Z(x), \label{eq2}
\end{equation}
where $\beta$ is a coefficient for the reverse KL divergence penalty, and $Z(x)$ is the partition function.

Given the chosen response $y_w$, preferred over the rejected response $y_l$, DPO aligns with human preference using the Bradley-Terry model for pairwise comparisons:
\setlength\abovedisplayskip{3pt}
\setlength\belowdisplayskip{3pt}
\begin{equation}
    P_{\mathrm{BT}}(y_w\succ y_l|x)=\frac{\exp(r(x,y_w))}{\exp(r(x,y_w))+\exp(r(x,y_l))}. \label{eq3}
\end{equation}
By substituting Eq.~\ref{eq2} into Eq.~\ref{eq3} and leveraging the negative log-likelihood loss, DPO derives the objective function: 
\setlength\abovedisplayskip{3pt}
\setlength\belowdisplayskip{3pt}
\begin{equation}
    \begin{aligned}
    u(x,y_w,y_l)=\beta\log\frac{\pi_\theta(y_w|x)}{\pi_{\mathrm{ref}}(y_w|x)}-\beta\log\frac{\pi_\theta(y_l|x)}{\pi_{\mathrm{ref}}(y_l|x)},\\ \mathcal{L}_{\mathcal{DPO}}=-\mathbb{E}_{(x,y_w,y_l)}\left[\log\sigma\left(u(x,y_w,y_l)\right)\right],
    \end{aligned}
\end{equation}
where the action score with $y_i$ denotes the $i$-th token of the response $y$ can be formulated as:
\setlength\abovedisplayskip{3pt}
\setlength\belowdisplayskip{3pt}
\begin{equation}
    \log\pi(y|x)=\sum_{y_i\in y}\log p(y_i|x,y_{<i}).
\end{equation}

\begin{figure*}[!t]
  \centering
   \includegraphics[width=0.98\linewidth]{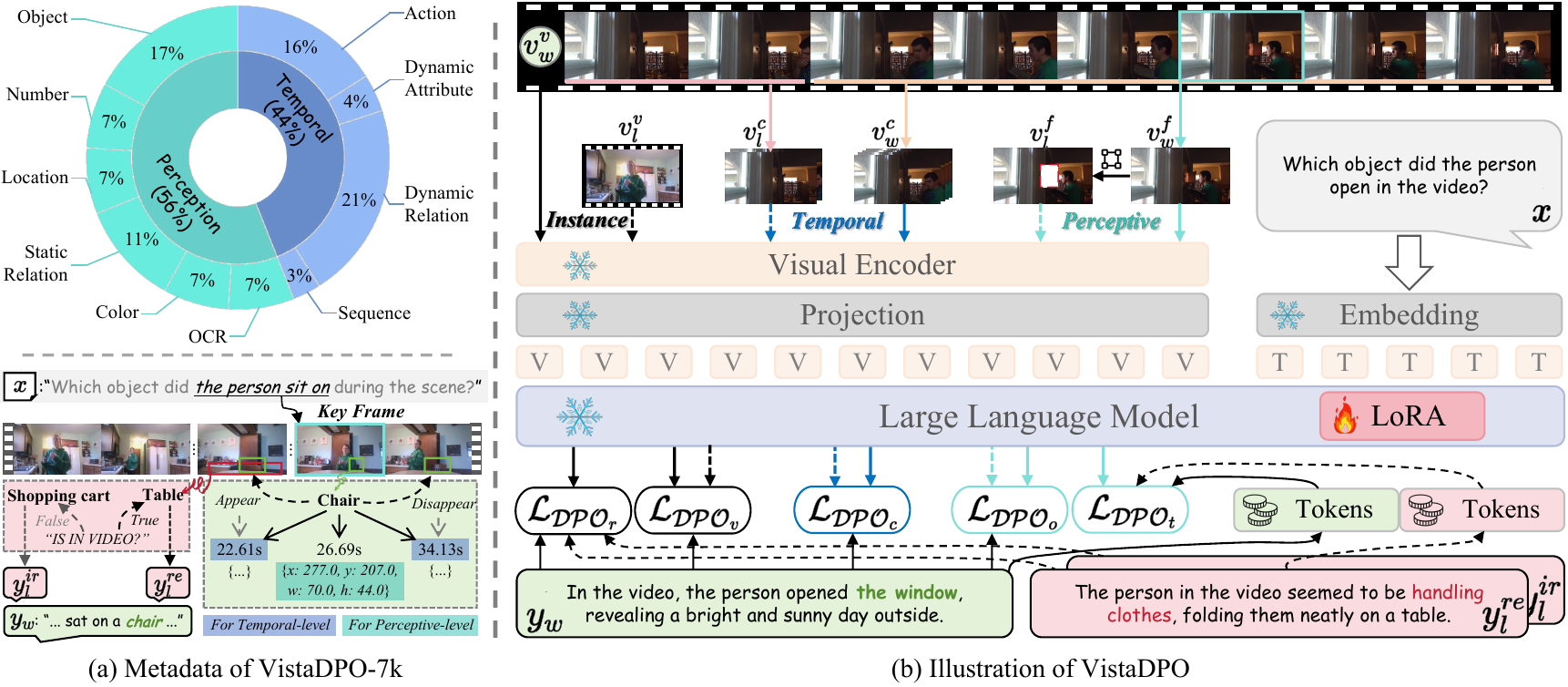}
   \vspace{-2mm}
   \caption{(a) The metadata of VistaDPO-7k highlights its focus on fine-grained video-language tasks, emphasizing temporal ($44\%$) and perceptual ($56\%$) reasoning. $y_l^{ir}$ and $y_l^{re}$ denote the irrelevant and relevant non-preferred responses respectively. (b) VistaDPO introduces a hierarchical spatiotemporal preference optimization framework. Instance ($v^v$) and perceptive ($v^f$) levels align global-to-local semantics with spatial visual features, leveraging both text-relevant and irrelevant rejected responses for robust cross-modal interaction. Temporal ($v^c$) level aligns clip-level semantics with temporal dynamics, enabling precise reasoning across spatial and temporal dimensions.}
   \vspace{-4mm}
   \label{fig:pipeline}
\end{figure*}

% Constructing 
\vspace{-2mm}
\section{\texttt{VistaDPO-7k}: A Spatial-temporal Grounded Video DPO Dataset}
\label{sec:dataset}
Existing LVMs often suffer from limited spatial-temporal perception, leading to video-language misalignment and hallucination issues~\cite{lan2024survey}. 
We propose VistaDPO with spatial-temporal DPO to achieve fine-grained alignment between video and language modalities. 
To support this, we construct a spatial-temporal grounded dataset, \textbf{\texttt{VistaDPO-7k}}, by integrating data from $14$ prevalent video datasets and systematically designing QA pairs to evaluate and mitigate hallucinations. 
These hallucinations are categorized into two major dimensions: Perception (\emph{e.g.}, Object, Static/Dynamic Attribute, Static Relation, OCR) and Temporal (\emph{e.g.}, Action, Dynamic Relation, Sequence), covering both static and dynamic aspects of video understanding. 
The dataset provides chosen and rejected responses, along with fine-grained temporal dependencies that include key timestamps, frames, and bounding boxes, enabling models to better capture spatial-temporal interactions, as can be shown in Figure~\ref{fig:pipeline}(a). 
\texttt{VistaDPO-7k} supports multi-level preference optimization across Temporal, Perceptive, and Instance levels, offering a robust benchmark to reduce hallucinations and enhance the spatial-temporal reasoning capabilities of LVMs.
Please refer to \textbf{Appendix} \S\ref{app_sec:data_pipeline} for more details on dataset construction and specifications.
\vspace{-0.4em}

% \begin{figure}[!t]
%   \centering
%    \includegraphics[width=0.98\linewidth]{Figures/placeholder.pdf}
%    \vspace{-2mm}
%    \caption{\scott{Data snapshot Placeholder} }
%    \vspace{-4mm}
%    \label{fig:data}
% \end{figure}

\vspace{-1mm}
\section{Methodology}
\label{sec:method}

\vspace{-1mm}
% \scott{Overview description of the system.}
To tackle the spatiotemporal complexities in video-language tasks, we propose VistaDPO, which implements hierarchical preference optimization across three aspects: (i) Instance-wise Semantic Preference Optimization, aligning preferences at response and video levels; (ii) Temporal Action-Event Preference Optimization, capturing overlooked temporal dynamics; and (iii) Perceptive Spatial-Object Preference Optimization, enabling fine-grained alignment between tokens and objects.
Figure~\ref{fig:pipeline}(b) illustrates the overall architecture of VistaDPO.

\vspace{-2.4mm}
\subsection{Instance-wise Semantic Preference Optimization}
\vspace{-1.6mm}
Effective video-language alignment hinges on distinguishing preferred (chosen) from non-preferred (rejected) responses while capturing global video content. To address hallucinations and misalignments caused by spatiotemporal complexities and over-reliance on text, we propose response-level alignment to refine preference differentiation and video-level alignment to enhance instance-wise semantic understanding.

\textbf{Response-Level Alignment.}
LVMs often face challenges in maintaining global consistency when generating responses. While these models effectively capture the general context of video input $v$ and prompt $x$, they frequently struggle to distinguish user-preferred responses $y_w$ from non-preferred responses $y_l$ at the response level, leading to suboptimal alignment with user intent.
% This issue typically stems from insufficient modeling of the spatiotemporal relationships and scene-level semantics inherent in video data.
To promote overall consistency by encouraging the model to align its response-level preferences with human expectations, the objective function can be formulated as:
\setlength\abovedisplayskip{3pt}
\setlength\belowdisplayskip{3pt}
\begin{equation}
    \mathcal{L}_{\mathcal{DPO}_r}=-\mathbb{E}_{(v,x,y_w,y_l)}\left[\log\sigma\left(u_r(v,x,y_w,y_l)\right)\right],
\end{equation}
where
\begin{equation}
    u_r=\beta\log\frac{\pi_\theta(y_w|v,x)}{\pi_{\mathrm{ref}}(y_w|v,x)}-\beta\log\frac{\pi_\theta(y_l|v,x)}{\pi_{\mathrm{ref}}(y_l|v,x)}.
\end{equation}
 Here, $\log\pi(y|v,x)$ is defined as: 
\setlength\abovedisplayskip{3pt}
\setlength\belowdisplayskip{3pt}
\begin{equation}
    \log\pi(y|v,x)=\sum_{y_i\in y}\log p(y_i|v,x,y_{<i}).
\end{equation}

The existing method of Hound-DPO~\cite{zhang2024direct} directly adopts the above approach, focusing solely on aligning the chosen response with the prompt. Nevertheless, the complex spatial-temporal dependencies in rejected responses are completely neglected. 
Intuitively, intrinsic hallucinations in generative models typically arise from: 1) erroneously inferring content that does not exist in the video; 2) failing to capture the fine-grained spatial-temporal dependencies of the correct content in the video. To mitigate this, we further introduce two types of non-preferred responses into the optimization process:
\setlength\abovedisplayskip{3pt}
\setlength\belowdisplayskip{3pt}
\begin{equation}
\begin{aligned}
    \log\frac{\pi_\theta(y_l|v,x)}{\pi_{\mathrm{ref}}(y_l|v,x)} \leftarrow \sum_{i \in \{re, ir\}} \beta_i \log\frac{\pi_\theta(y_l^i|v,x)}{\pi_{\mathrm{ref}}(y_l^i|v,x)},
\end{aligned}
\end{equation}
where $y_l^{re}$ denotes the \textit{relevant} non-preferred for these are semantically relevant to the video content but contain spatial or temporal inconsistencies, \textit{e.g.}, incorrect temporal ordering, wrong actions, or misinterpreted spatial locations. In contrast, $y_l^{ir}$ denotes the \textit{irrelevant} non-preferred responses, which are entirely unrelated to the video content, introducing noise by hallucinating events or objects with no connection to the actual video.

\textbf{Video-Level Alignment.}
Unlike most prior DPO works, which focus exclusively on textual optimization, we introduce video-level preference optimization for the first time to reduce LVMs' overreliance on language. At the video level, the model needs to understand the preference relationships of the entire video as a coherent semantic unit. However, since LVMs are prone to hallucinations involving irrelevant video content, we optimize the model to recognize global discrepancies among videos. To this end, we construct video-level preferred and non-preferred sample pairs, denoted as $v_w^v$ and $v_l^v$. Thus $u_v(v_w^v,v_l^v,x,y_w)$ within $\mathcal{L}_{\mathcal{DPO}_v}$ can be formulated as:
\setlength\abovedisplayskip{3pt}
\setlength\belowdisplayskip{3pt}
\begin{equation}
    u_v=\beta\log\frac{\pi_\theta(y_w|v_w^v,x)}{\pi_{\mathrm{ref}}(y_w|v_w^v,x)}-\beta\log\frac{\pi_\theta(y_l|v_l^v,x)}{\pi_{\mathrm{ref}}(y_l|v_l^v,x)},
    \label{eq9}
\end{equation}
where $v_l^v$ is sampled from the mini-batch that is unrelated to the query $x$ in this work.

\vspace{-2.4mm}
\subsection{Temporal Semantic Preference Optimization}
\vspace{-1.4mm}

\textbf{Clip-Level Alignment.}
While previous multimodal DPO methods have mainly focused on the spatial aspects of visual samples (as shown in Table~\ref{app_tab:dpo_comparison}), unlike static images, videos require both spatial semantic understanding and dynamic temporal reasoning. 
This necessitates a comprehensive modeling of the spatial-temporal attributes of videos.

At the temporal level, the model must distinguish between time segments in the video that are relevant to the prompt and those that are irrelevant. 
To align video temporal semantics with event descriptions provided in the prompt, we treat time segments related to the prompt as preferred clips $v_w^c$ and time segments unrelated to the prompt as non-preferred clips $v_l^c$, as shown in Figure~\ref{fig:pipeline}. 
Following Eq.~\eqref{eq9}, the clip-level objective function can be defined as:
\setlength\abovedisplayskip{3pt}
\setlength\belowdisplayskip{3pt}
\begin{equation}
    \mathcal{L}_{\mathcal{DPO}_c} \sim \log\sigma(u_c(v_w^c, v_l^c,x,y_w)). \label{eq10}
\end{equation}

\vspace{-2.4mm}
\subsection{Perceptive Spatial-Object Preference Optimization}
\vspace{-1.6mm}
% While higher-level alignment captures global semantics, fine-grained perceptual alignment is crucial for precise video-language interaction. Videos contain complex spatial relationships, where objects and actions interact within specific regions and time frames. Similarly, language refers to these objects and actions with specific tokens.

While instance-wise alignment captures global semantics, fine-grained perceptual alignment is crucial for precise video-language interaction. Videos inherently involve complex spatial relationships, where objects, actions, and regions dynamically interact over time. Language, in turn, encodes these interactions through specific tokens, making it essential to establish detailed alignment between spatial objects and their corresponding linguistic references. 
% This fine-grained alignment reduces hallucinations and ensures that the model captures the correct object-action relationships and temporal dependencies.

\textbf{Object-Level Spatial Alignment.}
At the spatial level, the model needs to capture the key locations and states of objects within the video. However, LVMs are often prone to hallucinations in spatial layouts, leading to incorrect object placements or misinterpretations of scene context. To address this, we strengthen the model's understanding of spatial information through object-level preferred and non-preferred sample design. Specifically, we select the keyframe relevant to the prompt $x$ as the preferred instance $v_w^f$ as shown in Figure.~\ref{fig:pipeline}. For the non-preferred sample $v_l^f$, we further apply a masking operation to the key regions within the selected frame, thereby focusing the model's attention on the relevant spatial content while reducing the influence of irrelevant regions. Accordingly, the object-level loss $\mathcal{L}_{\mathcal{DPO}_o}$ can be defined in a manner similar to Eq.~\eqref{eq10}.

\textbf{Token-Level Alignment.}
While response-level optimization enhances global consistency, it lacks the granularity required to address token-specific errors, such as misattributed objects or incorrect temporal markers (\textit{e.g.}, ``after" \textit{vs.} ``before"). Token-level optimization ensures that the model aligns its preferences at a finer granularity, thereby reducing hallucinations in object-action relationships. Inspired by TDPO~\cite{zeng2024token}, we implement token-level optimization to evaluate preferences for individual tokens and align them coherently to form a consistent response. The sequential KL divergence can be defined as:
\setlength\abovedisplayskip{3pt}
\setlength\belowdisplayskip{3pt}
\begin{equation}
\begin{aligned}
    \mathcal{L}_{DPO_t}=sg&\left(\beta D_\mathrm{SeqKL}(x,v_w^f,y_w;\pi_\mathrm{ref}||\pi_\theta)\right)\\&-\beta D_\mathrm{SeqKL}(x,v_w^f,y_l;\pi_\mathrm{ref}\|\pi_\theta),
\end{aligned}
\end{equation}
where $sg$ represents the stop-gradient operator, ensuring that gradients are not propagated through the reference policy $\pi_{\mathrm{ref}}$, and $D_{\mathrm{SeqKL}}$ is the sequence-level KL divergence:
\begin{equation}
    D_{\mathrm{SeqKL}}=\sum_{t=1}^TD_{\mathrm{KL}}(\pi_{\mathrm{ref}}(y|x,y_{<t})\|\pi_\theta(y|x,y_{<t})).
\end{equation}

% Finally, the \textit{Video Hierarchical Spatial-Temporal Direct Preference Optimization} includes instance, temporal, perceptive-level preference optimization. It can be formulated as:
Overall, after incorporating instance-wise, temporal, and perceptive-level preference optimization, the overall loss function for VistaDPO is formulated as follows:
\setlength\abovedisplayskip{3pt}
\setlength\belowdisplayskip{3pt}
\begin{equation}
\begin{aligned}
    \mathcal{L}_{VistaDPO}=&\underbrace{\mathcal{L}_{DPO_v}+\mathcal{L}_{DPO_r}}_{Instance}\\&+\underbrace{\lambda\mathcal{L}_{DPO_c}}_{\textcolor{NavyBlue}{Temporal}}+\underbrace{\mu\mathcal{L}_{DPO_o}+\rho\mathcal{L}_{DPO_t}}_{\textcolor{BlueGreen}{Perceptive}},
\end{aligned}
\end{equation}
where $\lambda$, $\mu$, and $\rho$ represent the loss weights.
\vspace{-0.6em}
% \subsection{Video-Language Co-view Preference Optimization}

% \lipsum[1]

\section{Experiments}
\label{sec:experiments}

In this section, we empirically investigate the effectiveness of VistaDPO in reducing hallucinations.

\vspace{-2.4mm}
\subsection{Experimental Settings}
\vspace{-1.2mm}

\begingroup
\begin{table*}[t]
\fontsize{8.5}{8}\selectfont 
\setlength{\tabcolsep}{0.8mm}
\vspace{-0.4em}
  \centering
  \caption{Main results on video \textit{hallucination} benchmarks. \textbf{Bold} values indicate the best performance and $\Delta$ denotes the corresponding improvement percentages over the baselines (\emph{i.e.} PLLaVA and Video-LLaVA).
  ``$\uparrow$'' denotes higher is better. 
  % $\Delta$ (\%) shows VistaDPO's improvement over the baseline (\emph{i.e.} PLLaVA and Video-LLaVA).
  }
  \centering
  % \vspace{-0.8em}
  % \footnotesize
   \begin{tabular}{l ccc ccccccc}
     % \hlineB{2.5}
     \toprule 
     % \hline
     \multirow{3}{*}{\textbf{Models}} & \multicolumn{3}{c}{\textbf{VideoHallucer}} & \multicolumn{7}{c}{\textbf{EventHallusion}}\\
     \cmidrule(lr){2-4}\cmidrule(lr){5-11}
     & \multirow{2}{*}{\textbf{Basic}$\uparrow$} & \multirow{2}{*}{\textbf{Hallucinated}$\uparrow$} & \multirow{2}{*}{\textbf{Overall}$\uparrow$} & \multicolumn{2}{c}{\textbf{Entire}} & \multicolumn{2}{c}{\textbf{Mix}} & \textbf{Misleading} & \multicolumn{2}{c}{\textbf{Overall}}\\
     \cmidrule(lr){5-6}\cmidrule(lr){7-8}\cmidrule(lr){9-9}\cmidrule(lr){10-11}
    % \cmidrule{5-6} \cmidrule{7-8} \cmidrule{9-9} \cmidrule{10-11}
     &  &  &  & \textbf{Binary}$\uparrow$ & \textbf{Desc.}$\uparrow$ & \textbf{Binary}$\uparrow$ & \textbf{Desc.}$\uparrow$ & \textbf{Binary}$\uparrow$ & \textbf{Binary}$\uparrow$ & \textbf{Desc.}$\uparrow$\\
     % \hlineB{2}
     \midrule
     VideoChatGPT~\cite{maaz2023video} & 92.8 & 10.4 & 6.4 & 14.9 & 5.5 & 57.0 & 3.6 & 21.6 & 36.4 & 4.3 \\
     VideoChat2~\cite{li2024mvbench} & 29.7 & 25.8 & 7.8 & 16.7 & 4.6 & 12.4 & 1.6 & 22.6 & 16.1 & 2.6 \\
     % PLLaVA~\cite{xu2024pllava} & 75.1 & 55.5 & 38.1 & 45.6 & 16.5 & 58.5 & 3.1 & 81.4 & 60.6 & 6.1 \\
     LLaMA-VID~\cite{li2025llama} & 89.9 & 26.6 & 21.0 & 30.7 & 16.5 & 73.6 & 7.8 & 43.1 & 54.0 & 10.9 \\
     \midrule
     PLLaVA~\cite{xu2024pllava} & 75.1 & 55.5 & 38.1 & 45.6 & 16.5 & 58.5 & 3.1 & 81.4 & 60.6 & 6.1 \\
     $+$ Hound-DPO~\cite{zhang2024direct} & 69.3  & 58.1 & 36.2 & 47.4 & 19.3 & 24.9 & 4.1 & 83.3 & 45.7 & 9.8 \\
     \rowcolor{cyan!10}
     $+$ \textbf{VistaDPO (Ours)} & \textbf{82.5} & \textbf{72.1} & \textbf{57.8} & \textbf{55.3} & \textbf{23.6} & \textbf{62.2} & \textbf{6.2} & \textbf{97.1} & \textbf{68.9} & \textbf{12.7} \\
     \rowcolor{cyan!10}
     $\Delta\%$ & 9.9 & 29.9 & 51.7 & 21.3 & 42.7 & 6.3 & 100.0 & 19.3 & 13.7 & 108.2 \\
     \midrule
     Video-LLaVA~\cite{lin2023video} & 95.1 & 20.3 & 17.8 & 30.7 & 8.3 & 57.5 & 7.3 & 41.2 & 45.9 & 7.6 \\
     $+$ Hound-DPO~\cite{zhang2024direct} &  83.4 & 43.0 & 29.5 & 35.9 & 9.8 & 15.5 & 9.3 & 63.7 & 33.3 & 9.5 \\
     % $+$ DPO~\cite{rafailov2024direct} &  - & - & - & - & - & - & - & - & - & - \\
     \rowcolor{cyan!10}
     $+$ \textbf{VistaDPO (Ours)} &  \textbf{98.2} & \textbf{64.4} & \textbf{54.3} & \textbf{50.9} & \textbf{14.9} & \textbf{62.2} & \textbf{10.4} & \textbf{95.1} & \textbf{67.2} & \textbf{12.1} \\
      \rowcolor{cyan!10}
      $\Delta\%$ & 3.3 & 217.2 & 205.1 & 65.8 & 79.5 & 8.2 & 42.5 & 130.8 & 46.4 & 59.2 \\
     % \hlineB{2.5}
     \bottomrule 
     % \hline
   \end{tabular}
  \label{tab:comparison_videohallu}
  \vspace{-0.2em}
\end{table*} 
\endgroup

\begingroup
\setlength{\tabcolsep}{2pt}
\begin{table*}[t]
\vspace{-1em}
\fontsize{8.5}{8}\selectfont 
\setlength{\tabcolsep}{1.mm}
  \centering
  \caption{Main results on video \textit{QA} and \textit{captioning} benchmarks. Symbols follow the definitions in Table~\ref{tab:comparison_videohallu}.}
  \centering
  % \vspace{-0.8em}
  % \scriptsize
  % \footnotesize
   \begin{tabular}{l cccc ccccc}
     % \hlineB{2.5}
     \toprule 
     % \hline
     \multirow{2}{*}{\textbf{Models}} & \multicolumn{4}{c}{\textbf{Question-Answer}} & \multicolumn{5}{c}{\textbf{Captioning}} \\
     \cmidrule(lr){2-5} \cmidrule(lr){6-10}
      & \textbf{MSVD}$\uparrow$ & \textbf{MSR-VTT}$\uparrow$ & \textbf{TGIF}$\uparrow$ & \textbf{Act.Net}$\uparrow$ & \textbf{Correct}$\uparrow$ & \textbf{Detail}$\uparrow$ & \textbf{Context}$\uparrow$ & \textbf{Temporal}$\uparrow$ & \textbf{Consist}$\uparrow$ \\
     % \hlineB{2}
     \midrule
     % VideoChat~\cite{li2023videochat} & 56.3 & 45.0 & 34.4 & - & 2.2 & 2.5 & 2.5 & 1.9 & 2.2 \\
     VideoChatGPT~\cite{maaz2023video} & 64.9 & 49.3 & 51.4 & 35.2 & 2.4 & 2.5 & 2.6 & 2.0 & 2.4\\
     LLaMA-Adapter~\cite{zhang2023llama} & 54.9 & 43.8 & - & 34.2 & 2.0 & 2.3 & 2.3 & 2.0 & 2.2 \\
     Video-LLaMA~\cite{zhang2023video} & 51.6 & 29.6 & - & 12.4 & 2.0 & 2.2 & 2.2 & 1.8 & 1.8\\
     \midrule
     PLLaVA~\cite{xu2024pllava} & 76.6 & 62.0 & 77.5 & 56.3 &  3.2 & 2.9 & 3.6 & 2.3 & 2.9 \\
     $+$ Hound-DPO~\cite{zhang2024direct} &  82.3 & 73.1 & 79.9 & 54.7 &  3.2 & 2.8 & 3.4 & 2.4 & 2.7  \\
     \rowcolor{cyan!10}
     $+$ \textbf{VistaDPO (Ours)} & \textbf{86.4} & \textbf{80.2} & \textbf{84.3} & \textbf{59.1} &  \textbf{3.5} & \textbf{3.0} & \textbf{3.9} & \textbf{2.8} & \textbf{2.9} \\
     \rowcolor{cyan!10}
     $\Delta\%$ & 12.8 & 29.4 & 8.8 & 5.0 &  9.4 & 3.5 & 8.3 & 21.7 & 0.0 \\
     \midrule
     Video-LLaVA~\cite{lin2023video} & 71.8 & 59.0 & 48.4 & 45.3 & 2.8 & 2.9 & 3.4 & 2.5 & 2.6\\
     $+$ Hound-DPO~\cite{zhang2024direct} &  80.7 & 70.2 & 61.4 & 40.9 &  3.0 & 2.7 & 3.3 & 2.0 & 2.6  \\
     \rowcolor{cyan!10}
     $+$ \textbf{VistaDPO (Ours)} & \textbf{85.3} & \textbf{76.9} & \textbf{74.1} & \textbf{55.0} & \textbf{3.4} & \textbf{2.9} & \textbf{3.6} & \textbf{2.6} & \textbf{2.9} \\
     \rowcolor{cyan!10}
     $\Delta\%$ & 18.8 & 30.3 & 53.1 & 21.5 & 21.4 & 0.0 & 5.9 & 4.0 & 11.5 \\
     % \hlineB{2.5}
     \hline 
     % \hline
   \end{tabular}
  \label{tab:comparison_videoqa_generative}
  \vspace{-1.2em}
\end{table*} 
\endgroup

\textbf{Baselines.}
We apply VistaDPO to two different 7B-size LVMs: Video-LLaVA~\cite{lin2023video} and PLLaVA \cite{xu2024pllava}. For Video-LLaVA, it employs LanguageBind~\cite{zhu2023languagebind} encoder for visual inputs, and Vicuna-7B v1.5~\cite{vicuna2023} as the LLM backbone. For PLLaVA, the visual input is processed through ViT-L \cite{radford2021learning} and MM projector, with Vicuna as the LLM backbone.
While other LVMs cannot be directly compared due to differences in base models, preference data, and alignment strategies, we provide these results for reference: VideoChatGPT \cite{maaz2023video}, VideoChat2 \cite{li2024mvbench}, LLaMA-VID \cite{li2025llama}, LLaMA-Adapter \cite{zhang2023llama}, and Video-LLaMA \cite{zhang2023video}.
% \lipsum[1]

\textbf{Evaluations.}
To evaluate the effectiveness of VistaDPO, we adopt benchmarks for three aspects: (1) \textit{Video Hallucination}: VideoHallucer \cite{wang2024videohallucer} and EventHallusion~\cite{zhang2024eventhallusion}; (2) \textit{General Video QA}: MSVD-QA \cite{xu2017video}, MSR-VTT-QA \cite{xu2017video}, TGIF-QA \cite{jang2017tgif}, and ActivityNet-QA \cite{yu2019activitynet}; and (3) \textit{Captioning Performance}: VideoChatGPT-Bench \cite{maaz2023video}. For ablation studies and analysis, we mainly employ our VistaDPO on Video-LLaVA.
% \lipsum[1]

\textbf{Implementation Details.}
We train the Video-LLaVA 7B \cite{lin2023video} and PLLaVA 7B \cite{xu2024pllava} with VistaDPO for 3 epochs, with a learning rate of $5e-7$ and a batch size of 8 on H100 GPUs. For training, we followed \citet{zhang2024direct} to set the hyperparameter $\beta=0.1$ and followed \citet{zeng2024token} to set $\rho=0.1$ for $\mathcal{L}_{DPO_t}$. As for hyperparameters of $\mathcal{L}_{DPO_c}$ and $\mathcal{L}_{DPO_o}$, we set $\lambda=0.4$ and $\mu=0.2$ respectively. Moreover, we set $\beta_{re}=0.7$ and $\beta_{ir}=0.3$ for the relevant and irrelevant non-preferred responses respectively for $\mathcal{L}_{DPO_r}$.

\vspace{-1.2mm}
\subsection{Main Results}
\vspace{-1.2mm}

% \begingroup
% \setlength{\tabcolsep}{3pt}
% \begin{table}[t]
% % \vspace{-0.4em}
% \renewcommand{\arraystretch}{1.1}
%   \centering
%   \caption{Main results on video \textit{generative} benchmarks. }

%   \centering
%   % \vspace{-0.8em}
%   \scriptsize
%    \begin{tabular}{l|ccccc}
%      % \hlineB{2.5}
%      \hline \hline
%      Models & Correctness & Detail & Context & Temporal & Consistency \\
%      \cline{2-5}
%      % \hlineB{2}
%      \hline
%      VideoChatGPT~\cite{maaz2023video} & 64.9 & 49.3 & 51.4 & 35.2 \\
%      Chat-UniVi~\cite{jin2024chat} & 65.0 & 54.6 & 60.3 & 45.8 \\
%      Video-LLaMA~\cite{zhang2023video} & 51.6 & 29.6 & - & 12.4 \\
%      \hline
%      VideoLLaVA~\cite{lin2023video} & 70.7 & 59.2 & 70.0 & 45.3 \\
%      $+$ Hound-DPO~\cite{zhang2024direct} &  - & - & - & -  \\
%      % $+$ DPO~\cite{rafailov2024direct} &  - & - & - & -  \\
%      \rowcolor{cyan!10}
%      $+$ \textbf{VistaDPO (Ours)} &  - & - & - & - \\
%      % \hlineB{2.5}
%      \hline \hline
%    \end{tabular}
%   \label{tab:comparison_videoqa}
%   % \vspace{-0.3cm}
% \end{table} 
% \endgroup
\begingroup
% \fontsize{8}{10}\selectfont 
% \setlength{\tabcolsep}{0.1mm}
% \setlength{\tabcolsep}{2.2pt}
\begin{table}[t]
\fontsize{8.5}{6.2}\selectfont 
\setlength{\tabcolsep}{2.2mm}
\vspace{-0.8em}
  \centering
  \caption{Ablation study of level losses on VideoHallucer. Hound-DPO \cite{zhang2024direct} employs the same strategy as DPO \cite{rafailov2024direct}, but based on its own constructed dataset.}
  \centering
  % \vspace{-0.8em}
  % \footnotesize
   \begin{tabular}{l ccc}
     % \hlineB{2.5}
     \toprule 
     % \hline
     \textbf{Methods} & \textbf{Basic}$\uparrow$ & \textbf{Hallu.}$\uparrow$ & \textbf{Over.}$\uparrow$\\
     % \midrule
     \midrule
     \rowcolor{cyan!10}
     VistaDPO & \textbf{98.2} & \textbf{64.4} & \textbf{54.3}  \\
     % \rowcolor{yellow!10}
     % $- \mathcal{L}_{DPO_v}$ & - & - & -  \\
     \rowcolor{yellow!10}
     w/o $\mathcal{L}_{DPO_c}$ & 97.8 & 62.3 & 53.0  \\
     \rowcolor{yellow!10}
     w/o $\mathcal{L}_{DPO_o}$ & 98.1 & 62.0 & 52.8  \\
     \rowcolor{yellow!10}
     w/o $\mathcal{L}_{DPO_o},\;\mathcal{L}_{DPO_t}$ & 97.6 & 61.5 & 49.4  \\
     \rowcolor{yellow!10}
     w/o $\mathcal{L}_{DPO_o},\;\mathcal{L}_{DPO_t},\;\mathcal{L}_{DPO_c}$ & 97.2 & 60.1 & 46.6  \\
     \rowcolor{yellow!10}
     only w/ $\mathcal{L}_{DPO_r}$ & 95.8 & 52.3 & 39.8  \\
     \midrule
     \rowcolor{yellow!10}
     Vanilla DPO w/ VistaDPO-7K & 95.4 & 50.8 & 38.1 \\
     Hound-DPO & 83.4 & 43.0 & 29.5 \\
     \bottomrule 
   \end{tabular}
  \label{tab:ablation_vista}
  \vspace{-5mm}
\end{table} 
\endgroup
We compare VistaDPO with Hound-DPO~\cite{zhang2024direct} on video hallucination, video QA, and captioning benchmarks to verify the effectiveness of our approach.

\textbf{Video Hallucination.} To benchmark VistaDPO, we focused on the model hallucination problem that DPO post-training aims to mitigate and compared its performance against the previous video DPO strategy, specifically Hound-DPO, based on LVMs PLLaVA \cite{xu2024pllava} and Video-LLaVA \cite{lin2023video}. As shown in Table~\ref{tab:comparison_videohallu}, we adopted two video hallucination benchmarks, VideoHallucer \cite{wang2024videohallucer} and EventHallusion \cite{zhang2024eventhallusion}. The results indicate that VistaDPO significantly alleviates hallucination issues compared to Hound-DPO. Notably, while Hound-DPO improved hallucination-related performance, they introduced undesirable trade-offs, such as reduced accuracy in addressing fundamental categories like the ``Basic" class in VideoHallucer. Furthermore, Hound-DPO led to a decline in the model's descriptive capabilities and accuracy, as observed in the ``Desc. (Descriptive)" category of EventHallusion. These limitations highlight the shortcomings of prior methods and underscore the superiority of our VistaDPO framework and the accompanying VistaDPO-7K dataset. To provide a comprehensive assessment of LVMs' performance post-training, we evaluate both their general and captioning capabilities in the following sections.

\begin{figure*}[t]
  \centering
   \includegraphics[width=1\linewidth]{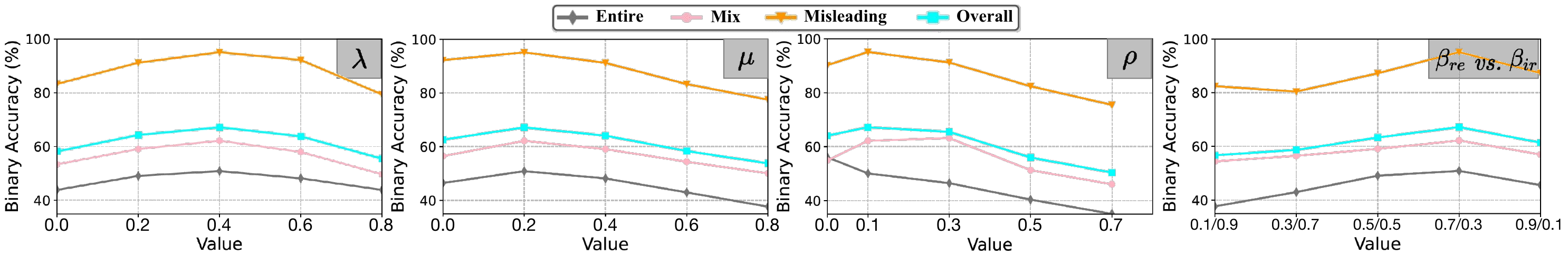}
   \vspace{-2em}
   \caption{Ablation study of hyperparameters on EventHallusion.}
   \vspace{-2mm}
   \label{fig:abla_hyper}
\end{figure*}

\begin{figure}[!t]
  \centering
   \includegraphics[width=1\linewidth]{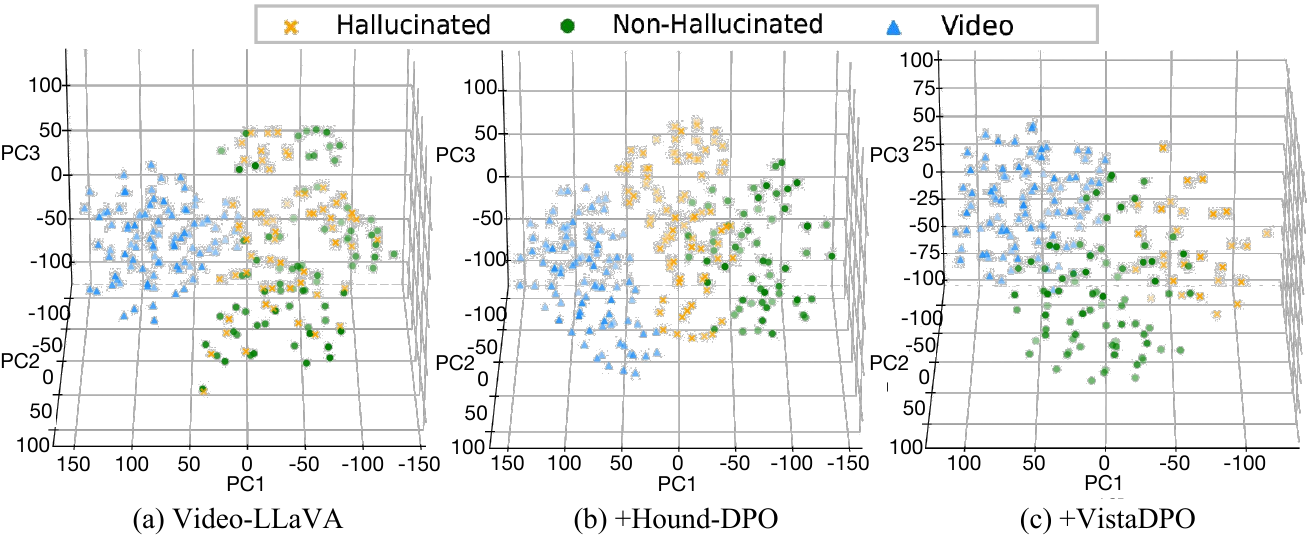}
   \vspace{-2em}
   \caption{T-SNE visualization of representation.~  
   (a) Video-LLaVA shows substantial overlap between hallucinated (orange) and non-hallucinated (green) representations. 
(b) With Hound-DPO, there is no distinct improvement in the separation of the two clusters.
(c) With VistaDPO, the representations achieve clear clustering, highlighting its superior discriminative capability.}
   \vspace{-6mm}
   \label{fig:tsne}
\end{figure}

\textbf{Video Question-Answering.} In addition to assessing the effectiveness of our VistaDPO in addressing hallucination issues, evaluating the model's general performance is equally critical. To this end, we conducted evaluations on four commonly used open-ended general question-answering benchmarks in a zero-shot setting, as illustrated on the left side of Table~\ref{tab:comparison_videoqa_generative}. VistaDPO consistently outperforms HoundDPO and demonstrates significant performance improvements on both base models. These results indicate that VistaDPO not only mitigates hallucination issues to a large extent but also enhances its ability to comprehend video content and generate accurate responses to questions.

\textbf{Captioning Capability.} We further evaluate the captioning capabilities of the model using the video-based text generation benchmark proposed by \citet{maaz2023video}, which assesses five critical dimensions: \underline{Correct}ness, \underline{Detail} Orientation, \underline{Context}ual Understanding, \underline{Temporal} Understanding, and \underline{Consist}ency. As shown on the right of Table~\ref{tab:comparison_videoqa_generative}, VistaDPO consistently outperforms Hound-DPO across all dimensions on two base models. These results highlight VistaDPO's ability to generate contextually relevant, detailed, and temporally accurate text from video inputs. Moreover, the findings demonstrate that the post-training process with VistaDPO-7K preserves the model's captioning capabilities, avoiding the degradation observed in Hound-DPO.

\vspace{-2.4mm}
\subsection{Ablation Studies}
\vspace{-1.2mm}
% All three levels—instance, temporal, and perceptive—significantly improve performance: (i) \textit{Instance}-level (VistaDPO-$\mathcal{L}_{DPO_o}$-$\mathcal{L}_{DPO_t}$-$\mathcal{L}_{DPO_c}$ \textit{vs.} DPO~\cite{rafailov2024direct}), (ii) \textit{Temporal}-level (VistaDPO \textit{vs.} VistaDPO-$\mathcal{L}_{DPO_c}$), and (iii) \textit{Perceptive}-level (VistaDPO \textit{vs.} VistaDPO-$\mathcal{L}_{DPO_o}$-$\mathcal{L}_{DPO_t}$). These suggest that the hierarchical granularity preference optimization allows the model to learn more effectively and align better.
To evaluate the contributions of each level and their combinations, we conduct ablation studies on VistaDPO using Video-LLaVA (Table~\ref{tab:ablation_vista}). The key findings are as follows:
\ding{182} \textbf{\textit{Effectiveness of Hierarchical Preference Optimization.}}
The hierarchical optimization strategy significantly improves performance, demonstrating its effectiveness in capturing multi-level preferences for better learning and task alignment.
\ding{183} \textbf{\textit{Importance of Spatial-Temporal Dependencies.}}
Spatial-temporal preference optimization, both explicit and implicit, plays a critical role in enhancing DPO performance: (i) VistaDPO explicitly captures spatial-temporal dependencies through object-level ($\mathcal{L}_{DPO_o}$) and clip-level ($\mathcal{L}_{DPO_c}$) optimization, enabling the model to better understand localized temporal and spatial relationships. (ii) Implicitly, it encodes spatial-temporal information via response-level ($\mathcal{L}_{DPO_r}$) preference alignment, which incorporates both relevant ($y_l^{re}$) and irrelevant ($y_l^{ir}$) non-preferred responses. These results highlight the importance of fine-grained spatial-temporal dependencies in video understanding, enabling more robust and effective video-language alignment.
\ding{184} \textbf{\textit{Impact of a Comprehensive High-quality Dataset.}}
Under the vanilla DPO strategy, post-training with VistaDPO-7K outperforms Hound-DPO, which uses a less comprehensive dataset. This demonstrates that a richer and higher-quality dataset improves generalization, enhances performance, and effectively mitigates hallucinations.  \ding{185} \textbf{Impact of Hyperparameters.} Additionally, we conduct hyperparameter ablation study (\emph{i.e.}~Figure~\ref{fig:abla_hyper}). Specifically, we analyzed the impact of two hyperparameter sets on VistaDPO performance: \ding{172} \textit{Loss Weights:} The optimal weights for all three levels balance the model's ability to capture temporal (clip-level $\lambda$), spatial (object-level $\mu$), and fine-grained token dependencies (token-level $\rho$). Too low a weight for any level weakens the model's ability to capture relevant dependencies, while excessively high weights disrupt the balance, leading to overfitting to specific details and loss of broader context. \ding{173} \textit{Weights for Relevant/Irrelevant Responses:} The combined weight for both non-preferred samples ($y_l^{re},\;y_l^{ir}$) helps the model capture spatial-temporal relationships at the textual level, which also highlights the need for careful hyperparameter tuning to effectively capture spatial-temporal relationships.

% \vspace{-1.4em}

% \textbf{Case Study of Different Losses.}

\vspace{-2mm}
\section{Analyses and Discussions}
\label{sec:analysis}

\vspace{-1mm}
We now take one step further, providing comprehensive analyses to demonstrate VistaDPO's superiority.
% via in-depth analyses.
% and robustness from representational and adversarial perspectives.

\begin{figure*}[h]
  \centering
   \includegraphics[width=1\linewidth]{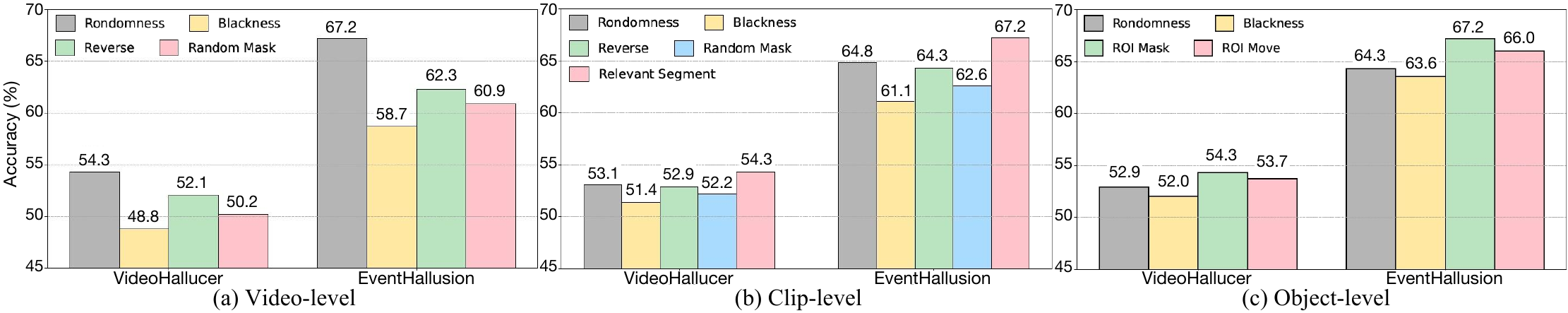}
   \vspace{-2.2em}
   \caption{Ablation study of visual non-preferred samples on two video hallucination benchmarks.}
   \vspace{-4mm}
   \label{fig:abla_visual_non}
\end{figure*}

\begin{figure}[!t]
  \centering
   \includegraphics[width=1\linewidth]{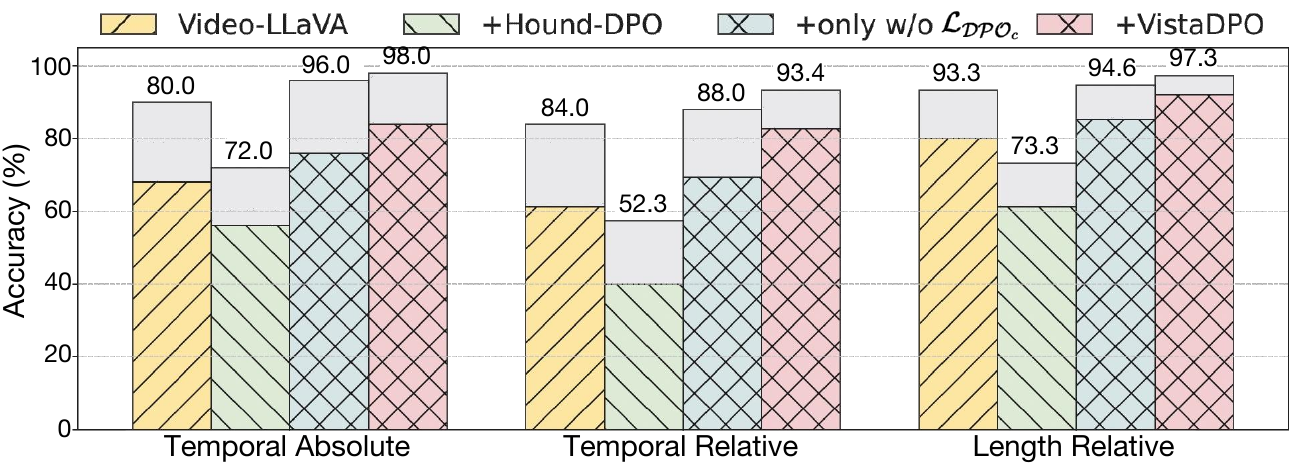}
   \vspace{-2.2em}
   \caption{Adversarial temporal testing on VideoHallucer. The gray regions indicate the performance drop under adversarial scenarios for each method. 
   % While Video-LLaVA and Hound-DPO show significant declines, VistaDPO exhibits minimal degradation, demonstrating its superior robustness in temporal modeling.
   }
   \vspace{-4mm}
   \label{fig:analysis_temporal}
\end{figure}

\vspace{-3mm}
% \subsection{Perspective-1: Enhanced Video-Language Representation}
\subsection{Enhanced Video-Language Representation}
\vspace{-1.2mm}
% Visualize Video-language alignment
To empirically demonstrate the effectiveness of VistaDPO, we conduct an analysis from a representational perspective, as illustrated in Figure~\ref{fig:tsne}. Specifically using $95$ samples (video, non-hallucinated captions, and hallucinated captions) from the ``misleading" subset of EventHallusion \cite{zhang2024eventhallusion}, we evaluated the alignment of visual and textual embeddings. Video-LLaVA exhibits overlapping features and weak modality alignment, struggling to distinguish hallucinated from non-hallucinated captions. With Hound-DPO, this issue is partially mitigated through vanilla DPO, but a significant gap between textual and video embeddings remains. In contrast, with VistaDPO, which incorporates hierarchical fine-grained preference modeling, the alignment is significantly improved by narrowing the distance between visual and textual modalities and distinctly separating hallucinated from non-hallucinated captions. These results underscore VistaDPO's superior capability to unify modalities and effectively reduce hallucination.

% Visual Negative Samples Analysis 
\vspace{-1em}
\subsection{Analysis of Visual Non-preferred Samples}
The quality of preference samples depends on the rejection visual samples and the gap between rejection and chosen samples. We explore strategies for constructing rejection samples at the video, clip, and object levels, while keeping the chosen samples (original video, event segment, and keyframe) unchanged for each level as shown in Figure~\ref{fig:abla_visual_non}.
\begin{compactitem}
    \item \textbf{Video-level:} (i) \textit{Randomness:} Select a random sample from the minibatch. (ii) \textit{Blackness:} Set all RGB values of the chosen sample to 0. (iii) \textit{Reverse:} Reverse the order of all frames in the chosen sample. (iv) \textit{Random Mask:} Mask half the frames in the chosen sample.
    \item \textbf{Clip-level:} (i) \textit{Randomness.} (ii) \textit{Blackness.} (iii) \textit{Reverse.} (iv) \textit{Random Mask.} (v) \textit{Relevant Segments:} Use segments where the event does not occur.
    \item \textbf{Object-level:} (i) \textit{Randomness.} (ii) \textit{Blackness.} (iii) \textit{ROI Mask:} Mask the key object in the chosen sample. (iv) \textit{ROI Move:} Move the key object to disrupt its original spatial relationships.
\end{compactitem}
As demonstrated in Figure~\ref{fig:abla_visual_non}, we observe the following performance trends:
Figure~\ref{fig:abla_visual_non} demonstrates the impact of different negative sample construction strategies across video, clip, and object levels on model performance. At the \textbf{video level}, the \textit{Reverse} method achieves the highest overall accuracy (67.2\%), significantly outperforming \textit{Randomness} (54.3\%), \textit{Blackness} (50.2\%), and \textit{Random Mask} (52.1\%). This suggests that disrupting temporal order provides more informative negative samples compared to random sampling or masking strategies \cite{chen2025temporal}, which fail to introduce sufficient semantic contrast. At the \textbf{clip level}, \textit{Relevant Segments} yields the best performance (64.8\%), surpassing \textit{Randomness} (53.1\%), \textit{Blackness} (52.9\%), \textit{Reverse} (61.1\%), and \textit{Random Mask} (62.6\%). This highlights that using event-irrelevant segments as negatives more effectively challenges the model to focus on event-specific semantics, whereas random or blackened clips lack meaningful contrast. At the \textbf{object level}, \textit{ROI Move} achieves the highest accuracy (66.0\%), outperforming \textit{ROI Mask} (64.3\%), \textit{Randomness} (54.3\%), and \textit{Blackness} (53.7\%). This indicates that spatially disrupting key objects introduces more challenging and informative negative samples compared to masking or random sampling. Overall, these results emphasize that well-designed, semantically targeted negative samples---such as those disrupting temporal order, leveraging event irrelevance, or altering spatial relationships---are crucial for enhancing the model's ability to distinguish fine-grained video-language alignments.

\begin{figure}[!t]
  \centering
   \includegraphics[width=1\linewidth]{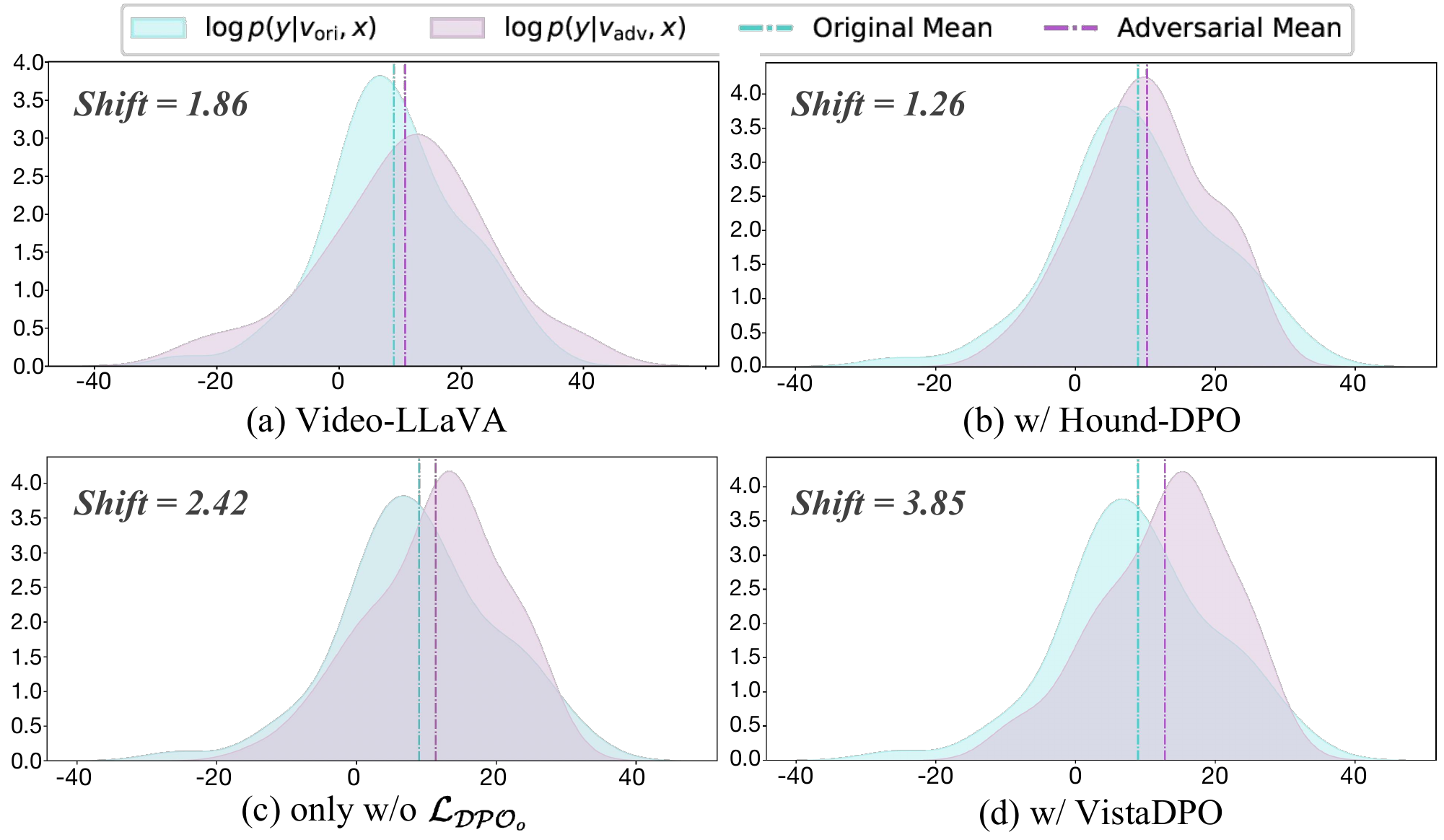}
   \vspace{-2.2em}
   \caption{Kernel Density Estimation (KDE) of log-likelihood differences in adversarial masking experiments. The log-likelihood difference measures the separation between original and adversarial distributions, with the shift representing the mean difference. Larger shifts indicate greater model robustness.}
   \vspace{-5mm}
   \label{fig:analysis_spatial_kde}
\end{figure}

\begin{figure*}[!t]
  \centering
  \vspace{-0.4em}
   \includegraphics[width=0.95\linewidth]{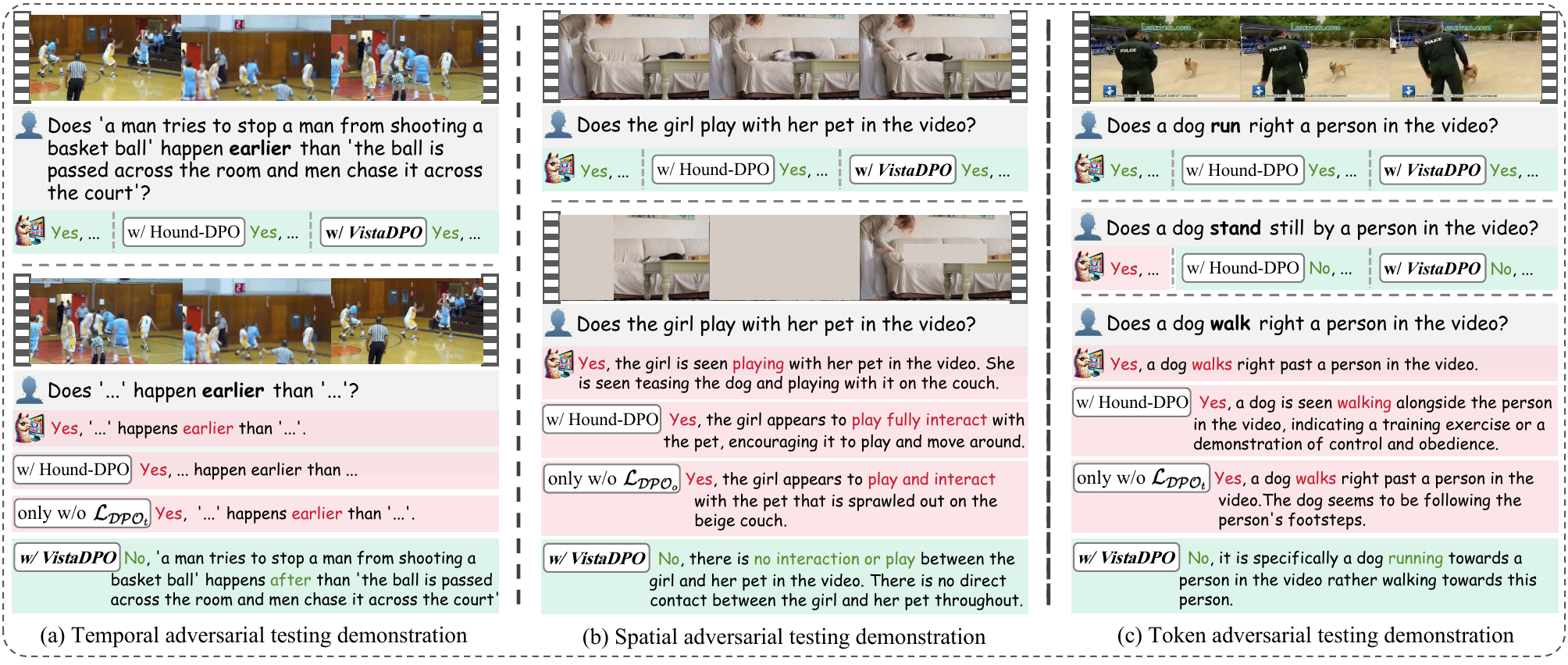}
   \vspace{-1.25em}
   \caption{Case Studies of Adversarial Testing for VistaDPO: We conduct case studies from three perspectives: (a) Temporal adversarial testing, which examines whether the model can infer the correct sequence of events by introducing reversed temporal order through video playback. (b) Spatial adversarial testing, which evaluates the model's ability to understand subject-object interactions by masking frames or pixels related to the target object. (c) Token adversarial testing, which tests the model's sensitivity to subtle linguistic differences by introducing similar action descriptions (e.g., contrasting ``run” with ``stand” and ``walk”). Each test compares VistaDPO with baselines (\emph{i.e.}, Video-LLaVA and Hound-DPO) and corresponding ablated versions to assess the impact of key components.}
   \vspace{-4mm}
   \label{fig:analysis_spatial}
\end{figure*}

% Adversarial Experiment
% \subsection{Analysis-2: Temporal Distorted Analysis (Clip)}

\vspace{-2.4mm}
% \subsection{Perspective-2: Adversarial Temporal Testing}
\subsection{Adversarial Temporal Testing}
\vspace{-1.2mm}
% Cases Study
To evaluate the robustness of VistaDPO, we conducted adversarial temporal testing using the ``Temporal” subset of VideoHallucer \cite{wang2024videohallucer}, which includes three categories of video-based QA tasks: (i) \textbf{Temporal Absolute}, focusing on when an event occurs; (ii) \textbf{Temporal Relative}, addressing the order of two events; and (iii) \textbf{Length Relative}, comparing the duration of two events. \textit{For adversarial testing, we reversed all videos and adjusted answers to align with the reversed timeline} (as shown in Figure~\ref{fig:analysis_spatial}(a). As shown in Figure~\ref{fig:analysis_temporal}, the base model (Video-LLaVA) and prior work (Hound-DPO) suffer significant performance drops across all three adversarial scenarios, revealing their inability to effectively model temporal hallucinations and vulnerability to timeline modifications. In contrast, VistaDPO shows minor degradation, demonstrating better temporal awareness and robustness against adversarial challenges.

% \subsection{Analysis-3: Spatial Distorted Analysis (Object)}
\vspace{-2.4mm}
% \subsection{Perspective-3: Adversarial Spatial Testing}
\subsection{Adversarial Spatial Testing}
\vspace{-1.2mm}
% Cases Study
% \textbf{Attention Visualization.}

To evaluate spatial adversarial robustness, we test with a video and the question, ``Does the girl play with her pet in the video?" As shown in Figure~\ref{fig:analysis_spatial}(b), all models correctly respond to the original video (upper side). However, in the adversarial version (lower side), where \textit{frames are masked to ensure the girl and pet never appear together}, only VistaDPO correctly identifies the absence of interaction. 
To further assess adversarial discriminative capability, we use Kernel Density Estimation (KDE) on the VideoHallucer dataset to visualize how model representations shift when reasoning over noisy (adversarial) samples (see Figure~\ref{fig:analysis_spatial_kde}). Video-LLaVA achieves a shift value of $1.86$, showing limited ability to distinguish between original and adversarial samples. Adding Hound-DPO slightly reduces the shift to $1.26$, indicating no improvement. VistaDPO achieves the highest shift value of $3.85$, significantly outperforming other models. Removing $\mathcal{L}_{DPO_o}$ reduces the shift to $2.42$, highlighting the importance of the proposed spatial-object preference optimization. These show VistaDPO's superior ability to capture subtle semantic differences and enhance adversarial robustness.

% This demonstrates VistaDPO's robust spatial reasoning, attributed to its object-level modeling, and highlights its resilience against adversarial perturbations by capturing fine-grained spatial relationships.

\vspace{-2.4mm}
% \subsection{Perspective-4: Adversarial Token Testing}
\subsection{Adversarial Token Testing}
\vspace{-1.2mm}

As shown in Figure~\ref{fig:analysis_spatial}(c), we conduct adversarial token testing to evaluate model robustness. 
For the original question, ``Does a dog run right a person in the video?'', all models answered correctly. When ``run'' was replaced with ``stand'' (\textit{a significant semantic shift}), most models maintained accurate responses. 
However, with an adversarial sample replacing ``run'' with ``walk'' (\textit{a subtle semantic change}), only VistaDPO correctly captured the nuanced difference. 
This underscores VistaDPO's robust token-level understanding, capturing fine-grained semantic shifts and ensuring precise video-language alignment under adversarial conditions.
\vspace{-0.6em}

\section{Conclusion}
\label{sec:conclusion}
\vspace{-1mm}

In this paper, we propose \textbf{VistaDPO}, a novel framework for Video Hierarchical Spatial-Temporal Direct Preference Optimization, which enhances the alignment between text and video preferences across three hierarchical levels: instance, temporal, and perceptive.
To support fine-grained preference alignment, we introduce \textbf{VistaDPO-7k}, a dataset of $7.2$K QA pairs with annotations for chosen/rejected responses and spatial-temporal groundings. 
Extensive evaluations on tasks, \emph{i.e.}, Video Hallucination, Video QA, and Captioning benchmarks demonstrate that VistaDPO significantly improves existing LVMs, addressing video-language misalignment and hallucination issues.
% Our method offers a promising step toward better multimodal alignment in video understanding.

% \textbf{Limitation}

\clearpage


% \section*{Acknowledgements}

% \textbf{Do not} include acknowledgements in the initial version of
% the paper submitted for blind review.

% If a paper is accepted, the final camera-ready version can (and
% usually should) include acknowledgements.  Such acknowledgements
% should be placed at the end of the section, in an unnumbered section
% that does not count towards the paper page limit. Typically, this will 
% include thanks to reviewers who gave useful comments, to colleagues 
% who contributed to the ideas, and to funding agencies and corporate 
% sponsors that provided financial support.

\section*{Impact Statement}
This paper presents work whose goal is to advance the field of Machine Learning, particularly in the domain of video-language alignment and large video models (LVMs). By introducing\textbf{\texttt{VistaDPO}}, a framework for hierarchical spatial-temporal direct preference optimization, and constructing the \textbf{\texttt{VistaDPO-7k}} dataset, we aim to improve the alignment between video content and human preferences, mitigating issues such as hallucination and misalignment in video-language tasks.

The potential societal impact of this work includes enhancing the robustness and reliability of AI systems in applications such as video analysis, autonomous systems, and multimedia content understanding. While these advancements could contribute positively to fields like education, accessibility, and entertainment, they also raise ethical considerations, including potential misuse in surveillance or biased decision-making if the models are not carefully evaluated for fairness and accountability.

We have taken steps to ensure that the dataset and methodology are designed to reduce biases and hallucinations, and we encourage future researchers to apply these methods responsibly. Beyond these considerations, there are no immediate societal consequences of this work that require specific attention.
% Authors are \textbf{required} to include a statement of the potential 
% broader impact of their work, including its ethical aspects and future 
% societal consequences. This statement should be in an unnumbered 
% section at the end of the paper (co-located with Acknowledgements -- 
% the two may appear in either order, but both must be before References), 
% and does not count toward the paper page limit. In many cases, where 
% the ethical impacts and expected societal implications are those that 
% are well established when advancing the field of Machine Learning, 
% substantial discussion is not required, and a simple statement such 
% as the following will suffice:

% ``This paper presents work whose goal is to advance the field of 
% Machine Learning. There are many potential societal consequences 
% of our work, none which we feel must be specifically highlighted here.''

% The above statement can be used verbatim in such cases, but we 
% encourage authors to think about whether there is content which does 
% warrant further discussion, as this statement will be apparent if the 
% paper is later flagged for ethics review.

\nocite{langley00}

\bibliography{example_paper}
\bibliographystyle{icml2025}

\newpage
\appendix
\onecolumn

\section{Limitation and Future Work}

While VistaDPO excels at aligning video and language with fine-grained precision, its performance on long-duration videos with complex temporal dependencies leaves room for improvement. Such scenarios pose unique challenges for any alignment framework. Building on our strong spatial-temporal modeling foundation, future work could explore hierarchical architectures or memory-augmented mechanisms to further enhance the ability to capture long-term interactions, extending the reach of our method to even more complex video-language tasks.

\section{More Details of Data Annotation}
\label{app_sec:data_pipeline}
\begin{table}[h]
\centering
\small % 调整字体大小以适应单栏
\vspace{-2mm}
\caption{Summary of Hallucination Types, Sample Counts, and Data Sources.}
\begin{tabular}{@{}p{4.5cm}p{2.5cm}p{9cm}@{}}
\toprule
\textbf{Hallucination Type} & \textbf{Sample Count} & \textbf{Data Source} \\ \midrule
Object & 1,200 & MSR-VTT, STAR, VATEX \\
Number & 500 & ActivityNet-QA, MSR-VTT, NExT-QA, VATEX \\
Location & 500 & MSR-VTT, NExT-QA, VATEX \\
Color & 500 & ActivityNet-QA, CLEVRER, MSR-VTT, VATEX \\
Static Relation & 800 & ActivityNet-QA, MSR-VTT, VATEX \\
OCR & 500 & RoadTextVQA, ViteVQA \\
Action & 1,200 & MSR-VTT, MSVD, STAR, VATEX \\
Dynamic Attribute & 300 & TempCompass, Tomato \\
Dynamic Relation & 1,500 & MSR-VTT, NExT-QA, STAR, VATEX, VCGBench-Diverse \\
Sequence & 200 & Video-MME, YouCook2 \\ \bottomrule
\end{tabular}
% \vspace{-2mm}
% \caption{Summary of Hallucination Types, Sample Counts, and Data Sources.}
\label{tab:hallucination_summary}
   \vspace{-1mm}
\end{table}
\paragraph{Datasets Sources.} We constructed a dataset by sampling from the validation sets of 14 existing datasets in Table~\ref{tab:hallucination_summary}, specifically MSR-VTT \cite{xu2016msr}, STAR \cite{mosig2020star}, VATEX \cite{wang2019vatex}, ActivityNet-QA \cite{yu2019activitynet}, NExT-QA \cite{xiao2021next}, CLEVRER \cite{yi2019clevrer}, RoadTextVQA \cite{tom2023reading}, ViteVQA \cite{zhao2022towards}, MSVD \cite{chen2011collecting}, TempCompass \cite{liu2024tempcompass}, Tomato \cite{shangguan2024tomato}, VCGBench-Diverse \cite{maaz2024videogpt+}, Video-MME \cite{fu2024video}, and YouCook2 \cite{ZhXuCoAAAI18}, encompassing tasks such as binary QA, multiple-choice QA, and captioning-QA. To define hallucination within the context of video-based QA, we categorized it into two dimensions: \textbf{Perception} and \textbf{Temporal}, and generated corresponding \textit{chosen} and \textit{rejected} responses.

\begin{figure}[!b]
  \centering
   \includegraphics[width=\linewidth]{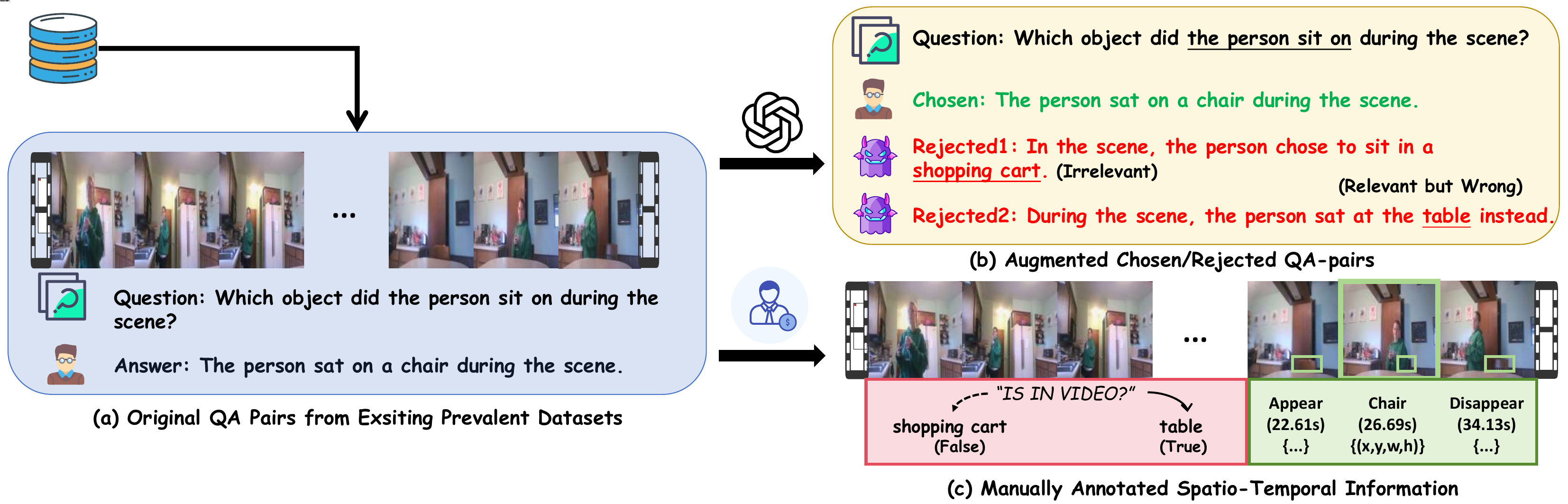}
   \vspace{-6mm}
   \caption{Illustration of dataset pipeline for constructing augmented video-language QA pairs. (a) Original QA pairs are extracted from existing prevalent datasets, providing basic QA pairs. (b) These pairs are augmented by introducing chosen and rejected answers, where rejected answers include both irrelevant responses (\textit{e.g.}, "shopping cart") and relevant but incorrect ones (\textit{e.g.}, "table"). (c) To enhance spatiotemporal understanding, manual annotations are added, specifying object appearances, spatial coordinates (\textit{e.g.}, bounding boxes), and temporal dynamics (\textit{e.g.}, appearance and disappearance timestamps). This pipeline ensures richer, more nuanced data for hierarchical preference optimization in video-language tasks.}
   \vspace{-4mm}
   \label{fig:data_pipeline}
\end{figure}

Specifically, the \textbf{Perception} dimension evaluates the model's ability to recognize static information in videos. This includes object recognition, identifying static attributes (\textit{e.g.}, number, color, position), understanding spatial relationships between objects, and extracting other elements such as OCR. In contrast, the \textbf{Temporal} dimension assesses the model's ability to comprehend dynamic temporal information, such as recognizing actions, identifying subtle dynamic attributes (\textit{e.g.}, movement direction, speed, shape), understanding event relationships, and perceiving action sequences within the video. By leveraging the prompt structure illustrated in Figure~\ref{fig:data_pipeline}, we expanded the original QA data into a dataset suitable for DPO training with \textit{chosen} and \textit{rejected} responses. During the construction of rejected response, we carefully considered whether the core semantics of the question were present in the video, generating both \textit{relevant} and \textit{irrelevant} rejected responses. This approach aims to enhance the model's global understanding and robustness at the response level.

To explicitly strengthen the model's spatiotemporal perception capabilities, we first identified all objects involved in the video. Subsequently, we manually annotated keyframes in which at least 30\% of the object's contours appeared or disappeared in the frame, as well as any keyframes directly relevant to answering the question. For each annotated keyframe, we labeled the bounding box coordinates (\emph{i.e.}, $(x,y,w,h)$) of the objects. 
\paragraph{Quality Control.}
To ensure annotation quality, all annotators were PhD students from universities who underwent standardized training and utilized a unified annotation tool. Each video was annotated independently by two annotators, and cross-validation was performed. Samples with annotation discrepancies were discarded to maintain high data quality.

\section{More Discussions on Related Work}

\begingroup
\begin{table}[h]
\fontsize{8}{10}\selectfont 
\setlength{\tabcolsep}{1.2mm}
% \vspace{-0.4em}
% \renewcommand{\arraystretch}{1}
  % \centering
  \vspace{-2mm}
  \centering
  \caption{Comparison among different DPO strategies.}
  % \vspace{-0.8em}
  % \scriptsize
   % \caption{Comparison among different DPO strategies.}
    % \vspace{-1em}
   % \vspace{-1em}
   \begin{tabular}{l c c ccc cc}
     \hline
     \multirow{2}{*}{\textbf{Method}} & \multirow{2}{*}{\textbf{LLM}} & \textbf{Base Model} & \multicolumn{3}{c}{\textbf{DPO}} & \textbf{Textual} & \textbf{Visual} \\
    \cline{4-6}
     % \cline{3-5}
     & & (7B, if not specified) & \textbf{Text} & \textbf{Image} & \textbf{Video} & \textbf{Granularity} & \textbf{Dimension} \\
     \hline 
     % \hline
     DPO~\cite{rafailov2024direct} & Text & Pythia-2.8B & \textcolor{green}{\ding{51}} & \textcolor{red}{\ding{55}} & \textcolor{red}{\ding{55}} & Sentence & \\
     IPO~\cite{azar2024general} & Text & Pythia-2.8B & \textcolor{green}{\ding{51}} & \textcolor{red}{\ding{55}} & \textcolor{red}{\ding{55}} & Sentence & \\
     KTO~\cite{ethayarajh2024kto} & Text & Llama-3-8B \& Qwen-3B-Instruct & \textcolor{green}{\ding{51}} & \textcolor{red}{\ding{55}} & \textcolor{red}{\ding{55}} & Sentence & \\
     R-DPO~\cite{park2024disentangling} & Text & Pythia-2.8B & \textcolor{green}{\ding{51}} & \textcolor{red}{\ding{55}} & \textcolor{red}{\ding{55}} & Sentence & \\
     SamPO~\cite{lu2024eliminating} & Text & Tulu2-13B-SFT \& Llama3-8B-Instruct & \textcolor{green}{\ding{51}} & \textcolor{red}{\ding{55}} & \textcolor{red}{\ding{55}} & Sentence & \\
     SePO~\cite{yang2024selective} & Text & LLaMA2-Chat \& Pythia-SFT-6.9B & \textcolor{green}{\ding{51}} & \textcolor{red}{\ding{55}} & \textcolor{red}{\ding{55}} & Sentence \& Token & \\
     TDPO~\cite{zeng2024token} & Text & GPT-2-Large& \textcolor{green}{\ding{51}} & \textcolor{red}{\ding{55}} & \textcolor{red}{\ding{55}} & Sentence \& Token & \\
     % Fr2Q~{\tiny\cite{rafailov2024from}} & Text & & \textcolor{green}{\ding{51}} & \textcolor{red}{\ding{55}} & \textcolor{red}{\ding{55}} & Sentence \& Token & \\
     \cdashline{1-8}
     HA-DPO~\cite{zhao2023beyond} & Image & LLaVA-v1.5 \& MiniGPT-4 & \textcolor{green}{\ding{51}} & \textcolor{red}{\ding{55}} & \textcolor{red}{\ding{55}} & Sentence & \\
     BPO~\cite{pi2025strengthening} & Image & LLaVA-v1.5 & \textcolor{green}{\ding{51}} & \textcolor{red}{\ding{55}} & \textcolor{red}{\ding{55}} & Sentence & \\
     FDPO~\cite{gunjal2024detecting} & Image & InstructBLIP-13B & \textcolor{green}{\ding{51}} & \textcolor{red}{\ding{55}} & \textcolor{red}{\ding{55}} & Sentence & \\
     HALVA~\cite{sarkar2024mitigating} & Image & LLaVA-v1.5 & \textcolor{green}{\ding{51}} & \textcolor{red}{\ding{55}} & \textcolor{red}{\ding{55}} & Sentence \& Token & \\
     POVID~\cite{zhou2024aligning} & Image & LLaVA-v1.5 & \textcolor{green}{\ding{51}} & \textcolor{green}{\ding{51}} & \textcolor{red}{\ding{55}} & Sentence & Spatial\\
     MIA-DPO~\cite{liu2024mia} & Image & LLaVA-v1.5 \& InternLM-XC2.5 & \textcolor{green}{\ding{51}} & \textcolor{green}{\ding{51}} & \textcolor{red}{\ding{55}} & Sentence & Spatial\\
     V-DPO~\cite{xie2024v} & Image & LLaVA-v1.5 & \textcolor{green}{\ding{51}} & \textcolor{green}{\ding{51}} & \textcolor{red}{\ding{55}} & Sentence & Spatial\\
     \cdashline{1-8}
     Next-DPO~\cite{li2024llava} & Video & LLaVA-Next & \textcolor{green}{\ding{51}} & \textcolor{red}{\ding{55}} & \textcolor{red}{\ding{55}} & Sentence & \\
     Hound-DPO~\cite{zhang2024direct} & Video & Video-LLaVA & \textcolor{green}{\ding{51}} & \textcolor{red}{\ding{55}} & \textcolor{red}{\ding{55}} & Sentence & \\
     \hline
     \textbf{VistaDPO (Ours)} &  Video & Video-LLaVA \& PLLaVA & \textcolor{green}{\ding{51}} &  \textcolor{green}{\ding{51}} & \textcolor{green}{\ding{51}} & Sentence \& Token & Spatial \& Temporal\\
     \hline
    %   \multicolumn{7}{l}{
    % \tiny$\diamondsuit$ Textual Granularity (\underline{S}entence, \underline{T}oken); $\nabla$ Visual Dimension (\underline{S}patial, \underline{T}emporal)
    % }
   \end{tabular}
  % \vspace{-0.4em}
  \label{app_tab:dpo_comparison}
\end{table} 
\endgroup

To highlight our contributions, we detail in Table~\ref{app_tab:dpo_comparison} how our proposed VistaDPO differs from previous DPO strategies. Two critical distinctions are summarized as follows:
\begin{compactitem}
    \item \textbf{Spatial-Temporal Video Preference Optimization:} Previous DPO methods predominantly focused on language-level alignment. However, with the ongoing development of multi-task~\cite{du2025dependeval,wu24next,maaz2024videogpt+} and multi-modal alignment~\cite{huang2024crest,huang2025trusted,wu24next,chen2024towards,huang2024evidential, yan2024urbanclip, luo2024nus, li2024survey, luo2024panosent}, reinforcement fine-tuning~\cite{tan2025reason,liu2025visual}, self-correction techniques~\cite{chen2024bovila}, and visual generation \cite{wu2025lightgen,chen2024gaussianvton,liu2024videodpo}, research efforts have gradually transitioned from pure language models towards vision-language models. While some works incorporated image-level visual alignment, these approaches remained limited to static images. Recent works like LLaVA-Next-DPO \cite{li2024llava} and LLaVA-Hound-DPO \cite{zhang2024direct} extended DPO strategies to video-language models. However, these methods only applied vanilla DPO strategies, optimizing alignment exclusively at the language level, with no explicit focus on visual modeling. In contrast, VistaDPO uniquely emphasizes optimizing spatial-temporal preferences in videos. By explicitly modeling both spatial and temporal preferences, VistaDPO bridges the gap between video content and textual understanding. This dual-layer spatial-temporal optimization enables our framework to address the complexities of video-language tasks comprehensively.
    \item \textbf{Hierarchical Finer Granularity:} Most existing DPO approaches operate at a coarse granularity, typically limited to sentence-level alignment for text and holistic-level alignment for visuals. With the development of fine-grained understanding and generation \cite{huang2025sefar, chen2024omnicreator}, advanced methods explore token-level textual alignment but still overlook hierarchical visual structures, which are crucial for video understanding. VistaDPO introduces a hierarchical granularity approach, incorporating both sentence- and token-level granularity for textual alignment and spatial- (object-) and temporal- (clip-) level granularity for visual alignment. By structuring alignment hierarchically across multiple layers—spanning from fine-grained token and object representations to coarse-grained sentence and video-level relationships—VistaDPO achieves a robust and precise preference optimization. This hierarchical approach empowers our framework to capture intricate cross-modal dependencies, ensuring superior performance in challenging scenarios such as adversarial testing and hallucination reduction.
\end{compactitem}

\section{Extended Details of Methodology: Formulas and Prompts}
This section details the core methodology used in VistaDPO, including the mathematical formulations and prompts employed during training. Key formulas for DPO are provided, along with the specific prompt templates used for generating and refining QA pairs. These details aim to provide a comprehensive understanding of the technical implementation.

\subsection{Formulations of Token-Level Preference Optimization.}
Token-Level Preference Optimization (TLPO) is a fine-grained optimization framework designed to align model outputs with human preferences by leveraging token-wise feedback. Unlike response-level optimization, TLPO avoids the cancellation of policies that may occur at the sentence level by focusing on sequential KL divergence at the token level.

\textbf{Human Preference Modeling.}
We employ the Bradley-Terry model to represent the probability of human preferences for a winning response \(y_w\) over a losing response \(y_l\), given the input \(x\) and auxiliary video context \(v_{w}^f\). The preference probability is defined as:
\[
P_{\text{BT}}(y_w \succ y_l | x, v_{w}^f) = \sigma\big(\lambda(x, v_{w}^f, y_w, y_l) - \delta(x, v_{w}^f, y_w, y_l)\big),
\]
where \(\sigma(\cdot)\) is the sigmoid function, \(\lambda(x, v_{w}^f, y_w, y_l)\) represents the difference in rewards, and \(\delta(x, v_{w}^f, y_w, y_l)\) is the difference in sequential KL divergence between the preference pairs. These terms are defined as follows:
\[
\lambda(x, v_{w}^f, y_w, y_l) = \beta \log \frac{\pi_\theta(y_w | x, v_{w}^f)}{\pi_\text{ref}(y_w | x, v_{w}^f)} - \beta \log \frac{\pi_\theta(y_l | x, v_{w}^f)}{\pi_\text{ref}(y_l | x, v_{w}^f)},
\]
\[
\delta(x, v_{w}^f, y_w, y_l) = \beta D_{\text{SeqKL}}(x, v_{w}^f, y_w; \pi_\text{ref} || \pi_\theta) - \beta D_{\text{SeqKL}}(x, v_{w}^f, y_l; \pi_\text{ref} || \pi_\theta).
\]

\textbf{Sequential KL Divergence.}
The sequential KL divergence \(D_{\text{SeqKL}}\) is defined as the sum of token-level KL divergences across the sequence:
\[
D_{\text{SeqKL}}(x, v_{w}^f, y; \pi_\text{ref} || \pi_\theta) = \sum_{t=1}^T D_{\text{KL}}(\pi_\text{ref}(y^t | x, v_{w}^f, y^{<t}) || \pi_\theta(y^t | x, v_{w}^f, y^{<t})),
\]
where \(T\) is the length of the sequence \(y\), and \(y^{<t}\) denotes the tokens generated up to step \(t-1\).

\textbf{Loss Function for TLPO.}
Combining the Bradley-Terry model and the sequential KL divergence, the loss function for TLPO is expressed as:
\[
\mathcal{L}_{\text{TLPO}} = -\mathbb{E}_{(x, v_{w}^f, y_w, y_l)} \left[ \log \sigma\big(\lambda(x, v_{w}^f, y_w, y_l) - \delta(x, v_{w}^f, y_w, y_l)\big) \right].
\]
Substituting \(\lambda(x, v_{w}^f, y_w, y_l)\) and \(\delta(x, v_{w}^f, y_w, y_l)\), the loss function can be rewritten as:
\begin{equation}
\begin{split}
\mathcal{L}_{\text{TLPO}} = 
& -\mathbb{E}_{(x, v_{w}^f, y_w, y_l)} \Bigg[ 
    \log \sigma \Bigg( 
        \beta \log \frac{\pi_\theta(y_w | x, v_{w}^f)}{\pi_\text{ref}(y_w | x, v_{w}^f)} 
        - \beta \log \frac{\pi_\theta(y_l | x, v_{w}^f)}{\pi_\text{ref}(y_l | x, v_{w}^f)} \\
& \quad - \alpha \big( 
        D_{\text{SeqKL}}(x, v_{w}^f, y_w; \pi_\text{ref} || \pi_\theta) 
        - \text{sg}(D_{\text{SeqKL}}(x, v_{w}^f, y_l; \pi_\text{ref} || \pi_\theta)) 
    \big) 
    \Bigg) 
\Bigg].
\end{split}
\label{eq:TLPO}
\end{equation}
where \(\alpha\) is a hyperparameter controlling the weight of the sequential KL divergence difference, and \(\text{sg}(\cdot)\) represents the stop-gradient operator.

\textbf{Final Formulation.}
The optimization term for TLPO, denoted as \(\mathcal{L}_{DPO_t}\), focuses solely on the sequential KL divergence difference:
\[
\mathcal{L}_{DPO_t} = \text{sg}\big(\beta D_{\text{SeqKL}}(x, v_{w}^f, y_w; \pi_\text{ref} || \pi_\theta)\big) - \beta D_{\text{SeqKL}}(x, v_{w}^f, y_l; \pi_\text{ref} || \pi_\theta).
\]

This term ensures that the learned policy \(\pi_\theta\) aligns closely with the winning sequence \(y_w\) while diverging from the losing sequence \(y_l\), effectively capturing human preferences at the token level.

\subsection{Prompt templates for Generating QA pairs.}
\begin{figure*}[h]
  \centering
   \includegraphics[width=0.75\linewidth]{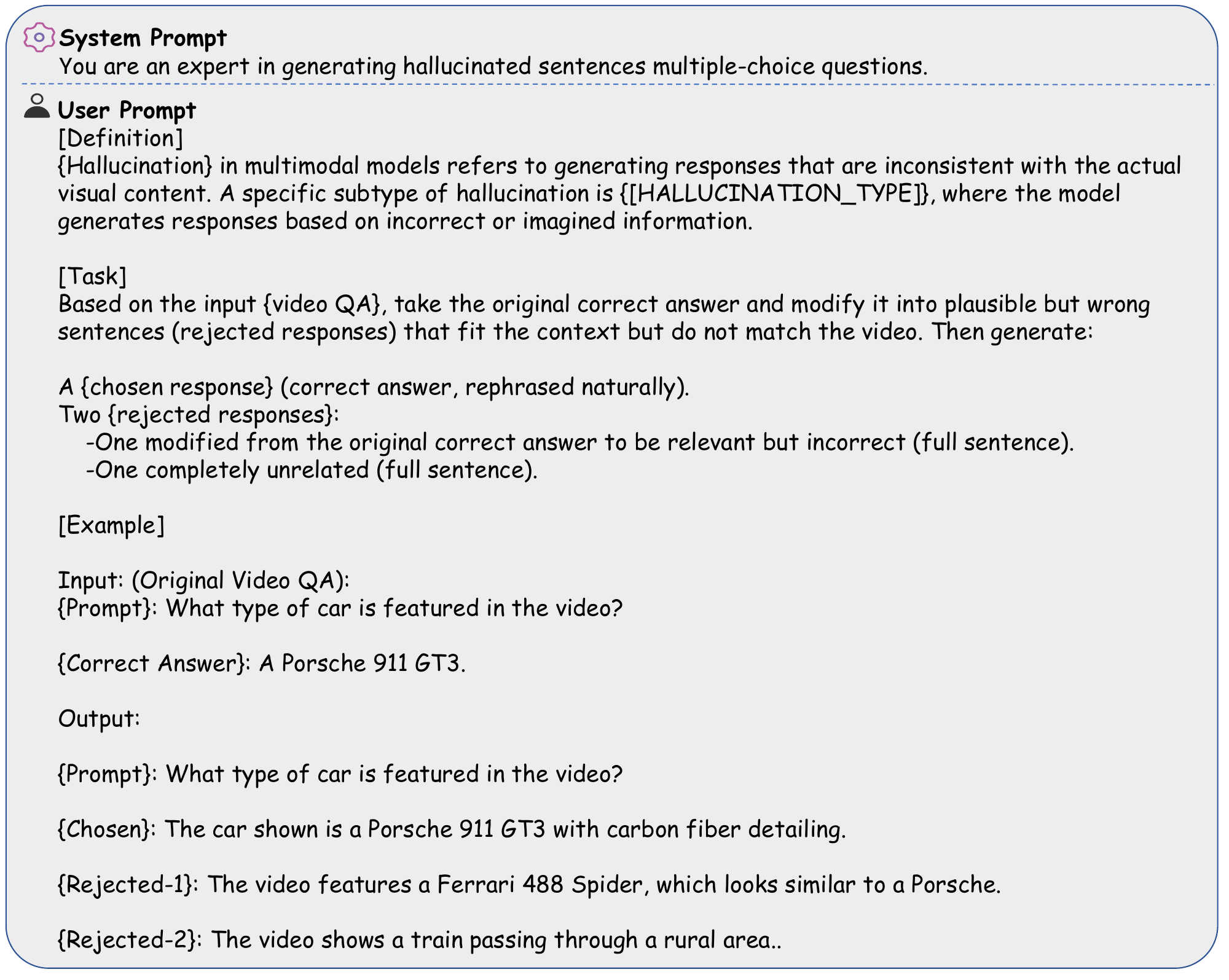}
   \vspace{-2mm}
   \caption{A prompt template designed for generating hallucinated responses in multimodal models is presented. The template transforms original video QA pairs into a "chosen response" (a rephrased correct answer) and two "rejected responses" (one contextually relevant but incorrect, and one entirely unrelated). This framework supports preference optimization by providing plausible yet inaccurate alternatives for training and evaluation. An example illustrates the process, highlighting the generation of both coherent and unrelated hallucinated responses.}
   \vspace{-4mm}
   \label{fig:prompt}
\end{figure*}

To adapt the existing dataset for fine-grained DPO training, we employed a template-based approach, as illustrated in Figure~\ref{fig:prompt}, and processed it using GPT-4. Specifically, we demonstrate the details of the prompt design using a multiple-choice dataset as an example.

\begingroup
\setlength{\tabcolsep}{2pt}
\begin{table*}[h]
% \vspace{-1em}
\fontsize{7}{7.5}\selectfont 
\setlength{\tabcolsep}{1.mm}
% \renewcommand{\arraystretch}{1.1}
  % \centering
  \vspace{-2mm}
  \centering
  \caption{Comparisons on MVBench. \textbf{Bold} values indicate the best performance achieved on the corresponding base model, while \underline{underlined} values represent the second-best performance. The results of VideoChat, VideoChatGPT, Video-LLaMA, and VideoChat2 are included as references, but they are not directly related to the contributions of this paper.}
  % \vspace{-0.8em}
  % \scriptsize
  % \footnotesize
   \begin{tabular}{l ccccccccccccccccccccc}
     % \hlineB{2.5}
     \toprule 
     % \hline
     \textbf{Models} & \textbf{Avg.} & \textbf{AS} & \textbf{AP} & \textbf{AA} & \textbf{FA} & \textbf{UA} & \textbf{OE} & \textbf{OI} & \textbf{OS} & \textbf{MD} & \textbf{AL} & \textbf{ST} & \textbf{AC} & \textbf{MC} & \textbf{MA} & \textbf{SC} & \textbf{FP} & \textbf{CO} & \textbf{EN} & \textbf{ER} & \textbf{CI} \\
     \midrule
     VideoChat~\cite{li2023videochat} & 35.5 & 33.5 & 26.5 & 56.0 & 33.5 & 40.5 & 53.0 & 40.5 & 30.0 & 25.5 & 27.0 & 48.5 & 35.0 & 20.5 & 42.5 & 46.0 & 26.5 & 41.0 & 23.5 & 23.5 & 36.0  \\
     VideoChatGPT~\cite{maaz2023video} & 32.7 & 23.5 & 26.0 & 62.0 & 22.5 & 26.5 & 54.0 & 28.0 & 40.0 & 23.0 & 20.0 & 31.0 & 30.5 & 25.5 & 39.5 & 48.5 & 29.0 & 33.0 & 29.5 & 26.0 & 35.5 \\
     Video-LLaMA~\cite{zhang2023video} & 34.1 & 27.5 & 25.5 & 51.0 & 29.0 & 39.0 & 48.0 & 40.5 & 38.0 & 22.5 & 22.5 & 43.0 & 34.0 & 22.5 & 32.5 & 45.5 & 32.5 & 40.0 & 30.0 & 21.0 & 37.0\\
     VideoChat2~\cite{li2024mvbench} & 51.1 & 66.0 & 47.5 & 83.5 & 49.5 & 60.0 & 58.0 & 71.5 & 42.5 & 23.0 & 23.0 & 88.5 & 39.0 & 42.0 & 58.5 & 44.0 & 49.0 & 36.5 & 35.0 & 40.5 & 65.5 \\
     \midrule
     PLLaVA~\cite{xu2024pllava} & \underline{46.6} & \underline{58.0} & \underline{49.0} & 55.5 & \underline{41.0} & \textbf{61.0} & \underline{56.0} & 61.0 & \textbf{36.0} & \underline{23.5} & 26.0 & 82.0 & \underline{39.5} & \underline{42.0} & \textbf{52.0} & \underline{45.0} & \underline{42.0} & \underline{53.5} & 30.5 & \underline{48.0} & 31.0 \\
     $+$ Hound-DPO~\cite{zhang2024direct} & 45.3 & 54.0 & 46.0 & \underline{57.0} & 37.5 & \underline{59.5} & 54.5 & \underline{62.0} & 31.5 & \underline{23.5} & \underline{26.5} & \textbf{83.5} & 38.0 & 41.5 & 50.0 & 41.0 & 39.5 & 50.5 & \textbf{32.0} & 46.0 & \underline{32.5} \\
     \rowcolor{cyan!10}
     $+$ \textbf{VistaDPO (Ours)} & \textbf{49.3} & \textbf{59.5} & \textbf{51.0} & \textbf{60.0} & \textbf{41.5} & 59.0 & \textbf{64.5} & \textbf{66.0} & \underline{35.0} & \textbf{27.0} & \textbf{35.5} & \underline{82.5} & \textbf{40.0} & \textbf{45.5} & \underline{51.5} & \textbf{48.0} & \textbf{48.5} & \textbf{54.0} & \underline{31.0} & \textbf{50.0} & \textbf{35.0} \\
     \midrule
     Video-LLaVA~\cite{lin2023video} & 43.0 & \underline{46.0} & \underline{42.5} & 56.5 & \underline{39.0} & \textbf{53.5} & 53.0 & 48.0 & \textbf{41.0} & 29.0 & 31.5 & \textbf{82.5} & \underline{45.0} & 26.0 & \underline{53.0} & 41.5 & \underline{33.5} & 41.5 & 27.5 & 38.5 & 31.5 \\
     $+$ Hound-DPO~\cite{zhang2024direct} & \underline{43.3} & 44.5 & 40.0 & \textbf{59.0} & \underline{39.0} & \underline{52.5} & \underline{53.5} & \underline{49.5} & 36.5 & \underline{32.0} & \underline{33.5} & \underline{79.0} & 43.0 & \underline{28.0} & \textbf{55.5} & \underline{42.0} & 30.0 & \underline{43.0} & \textbf{31.0} & \underline{39.0} & \underline{35.0} \\
     \rowcolor{cyan!10}
     $+$ \textbf{VistaDPO (Ours)} & \textbf{46.3} & \textbf{47.5} & \textbf{45.0} & \underline{58.5} & \textbf{42.0} & 51.5 & \textbf{60.5} & \textbf{54.5} & \underline{39.5} & \textbf{36.0} & \textbf{37.5} & \textbf{82.5} & \textbf{49.0} & \textbf{28.5} & 51.0 & \textbf{49.0} & \textbf{39.5} &\textbf{ 44.0} & \underline{29.0} & \textbf{42.0} & \textbf{38.5} \\
     % \hlineB{2.5}
     \bottomrule 
     % \hline
   \end{tabular}
   % \caption{Comparisons on MVBench. \textit{Action}: Action Sequence (AS), Action Prediction (AP), Action Antonym (AA), Fine-grained Action (FA), Unexpected Action (UA); \textit{Object}: Object Existence (OE), Object Interaction (OI), Object Shuffle (OS); \textit{Position}: Moving Direction (MD), Action Localization (AL); \textit{Scene}: Scene Transition (ST); \textit{Count}: Action Count (AC), Moving Count (MC); \textit{Attribute}: Moving Attribute (MA), State Change (SC); \textit{Pose}: Fine-grained Pose (FP); \textit{Character}: Character Order (CO); \textit{Cognition}: Egocentric Navigation (EN), Episodic Reasoning (ER), Counterfactual Inference (CI).}
\parbox{\textwidth}{
{\small \textbf{Note: }\textit{Action}: Action Sequence (AS), Action Prediction (AP), Action Antonym (AA), Fine-grained Action (FA), Unexpected Action (UA); 
\textit{Object}: Object Existence (OE), Object Interaction (OI), Object Shuffle (OS); 
\textit{Position}: Moving Direction (MD), Action Localization (AL); 
\textit{Scene}: Scene Transition (ST); 
\textit{Count}: Action Count (AC), Moving Count (MC); 
\textit{Attribute}: Moving Attribute (MA), State Change (SC); 
\textit{Pose}: Fine-grained Pose (FP); 
\textit{Character}: Character Order (CO); 
\textit{Cognition}: Egocentric Navigation (EN), Episodic Reasoning (ER), Counterfactual Inference (CI).}
}
  \label{app_tab:comparison_mvbench}
  \vspace{-1.2em}
\end{table*} 
\endgroup

% \section{More Experimental Settings and Results, Visualizations}
\section{More Comparison on MVBench}
\label{sec:more_res}
% This section presents additional experimental details, results, and visualizations. It includes training details, comparisons, and ablation studies for VistaDPO. 
% Furthermore, detailed visualizations of spatial-temporal grounding annotations and model outputs are provided to illustrate the effectiveness of the proposed approach. Comparative results with baseline methods are also included to highlight performance improvements.

% \subsection{More Experimental Settings}

% \textbf{Benchmarks and Metrics}

% \subsection{Extended Training Details}

% xxxxxxxxxxxxx

% \subsection{More Comparison on MVBench}

To more comprehensively evaluate VistaDPO, we conduct tests on MVBench \cite{li2024mvbench}, which contains $4,000$ QA pairs across $11$ video datasets covering a wide range of scenes, ranging from first-person to third-person and from indoor to outdoor environments. These tasks are categorized into $20$ fine-grained temporal understanding tasks. The results in Table~\ref{app_tab:comparison_mvbench} shown an overall improvement of $2.7\%$ and $3.3\%$ compared to base model PLLaVA and Video-LLaVA, respectively. Notably, VistaDPO excels in Object Existence ($8.5\%$ and $7.5\%$), Object Interaction ($5.0\%$ and $6.5\%$), Moving Direction ($2.5\%$ and $7.0\%$), Action Localization ($9.5\%$ and $6.0\%$), and Fine-grained Pose ($6.5\%$ and $6.0\%$), demonstrating the effectiveness of our spatial-temporal and fine-grained modeling approach.

\section{Exhibition Board}

\begin{figure}[h]
  \centering
   \includegraphics[width=1\linewidth]{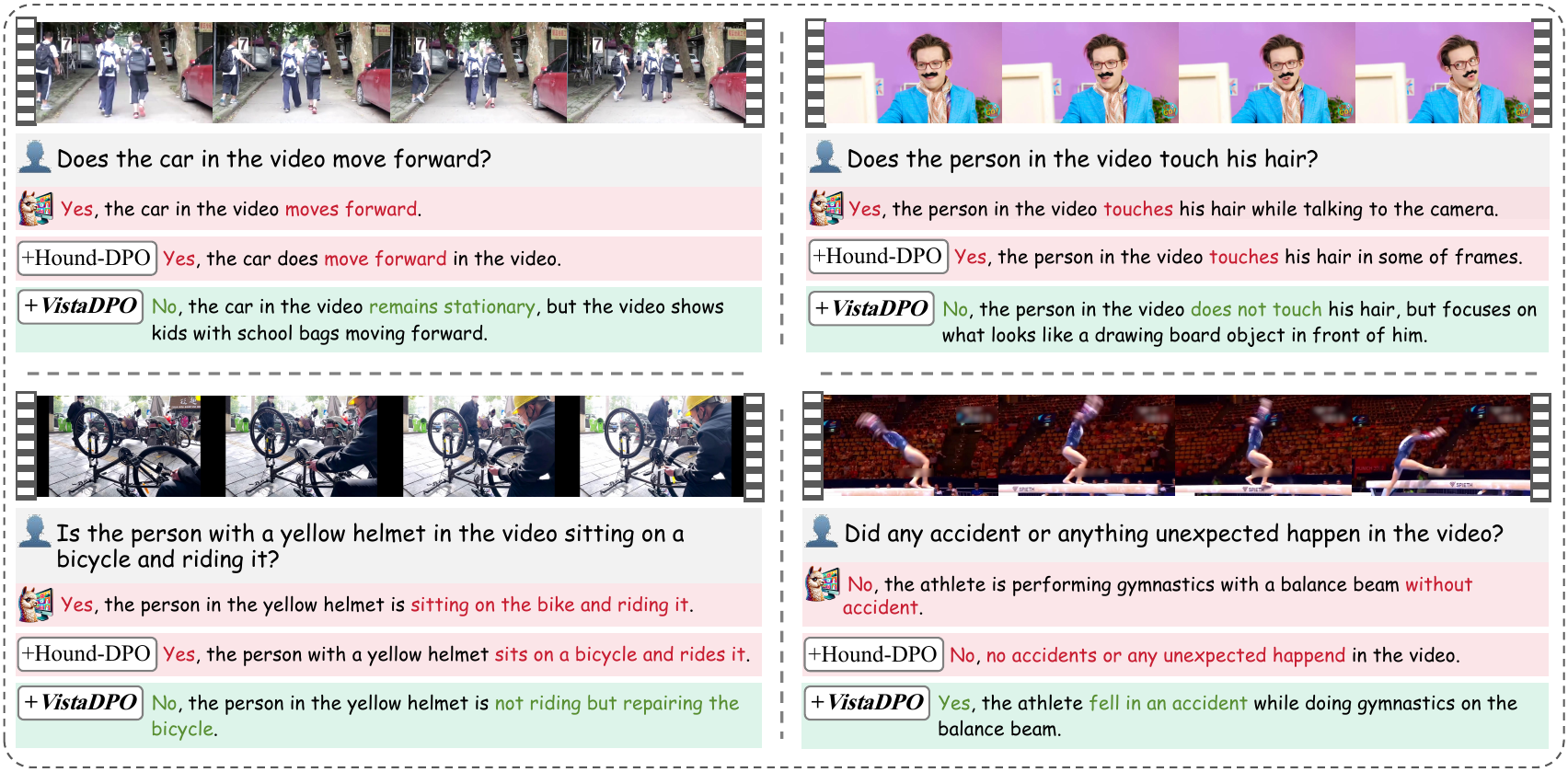}
   \vspace{-2em}
   \caption{Cases of VistaDPO in video understanding.}
   \vspace{-2mm}
   \label{app_fig:case}
\end{figure}

\textbf{Qualitative Demonstration.}
We show some unselected video QA cases in Figure \ref{app_fig:case}, which are sourced from VideoHallucer \cite{wang2024videohallucer} and EventHallusion \cite{zhang2024eventhallusion}.

\begin{figure}[h]
  \centering
   \includegraphics[width=0.8\linewidth]{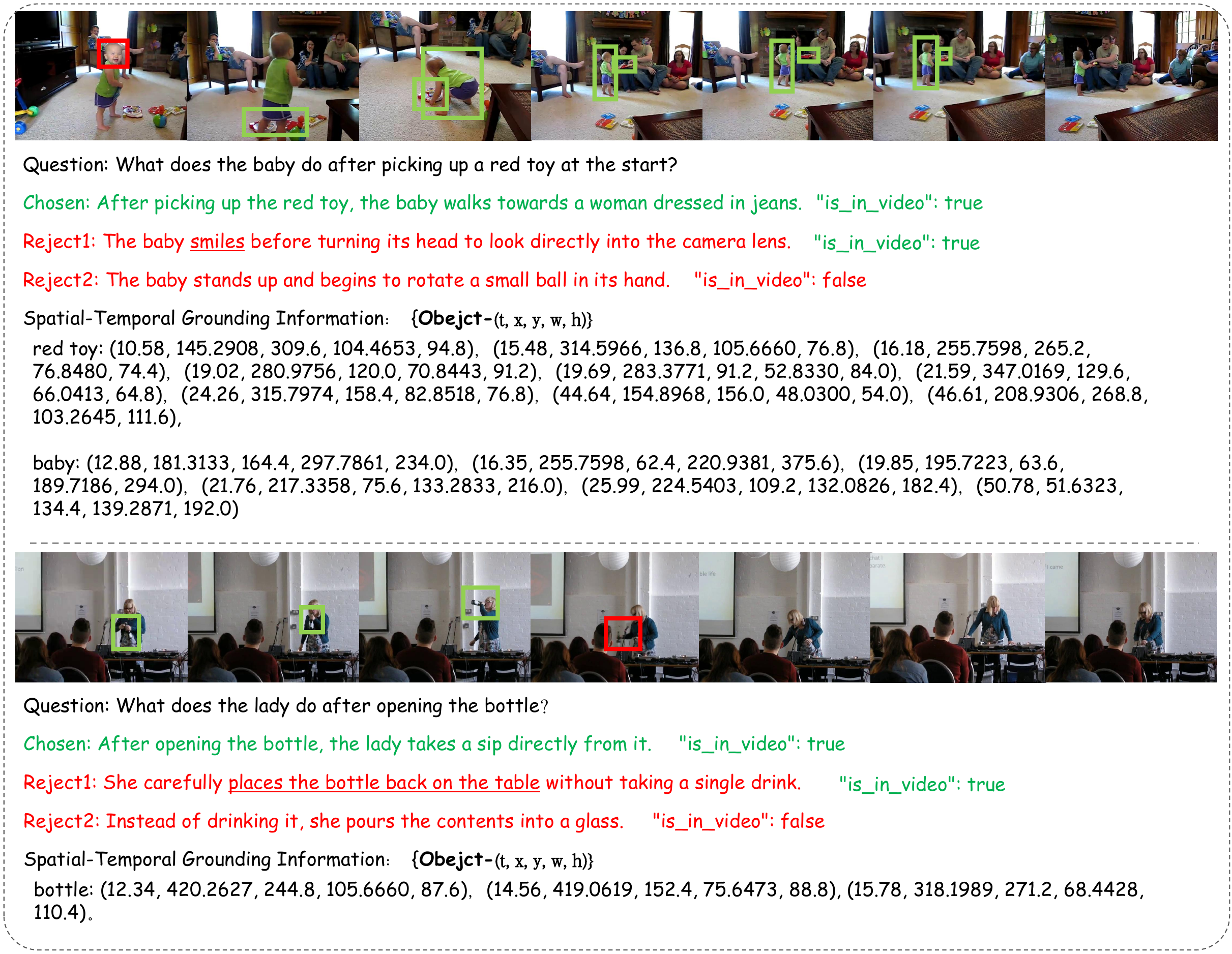}
   \vspace{-1em}
   \caption{Temporal data samples of VistaDPO-7K.}
   \vspace{-2mm}
   \label{app_fig:temp_data}
\end{figure}

\begin{figure}[h]
  \centering
   \includegraphics[width=0.8\linewidth]{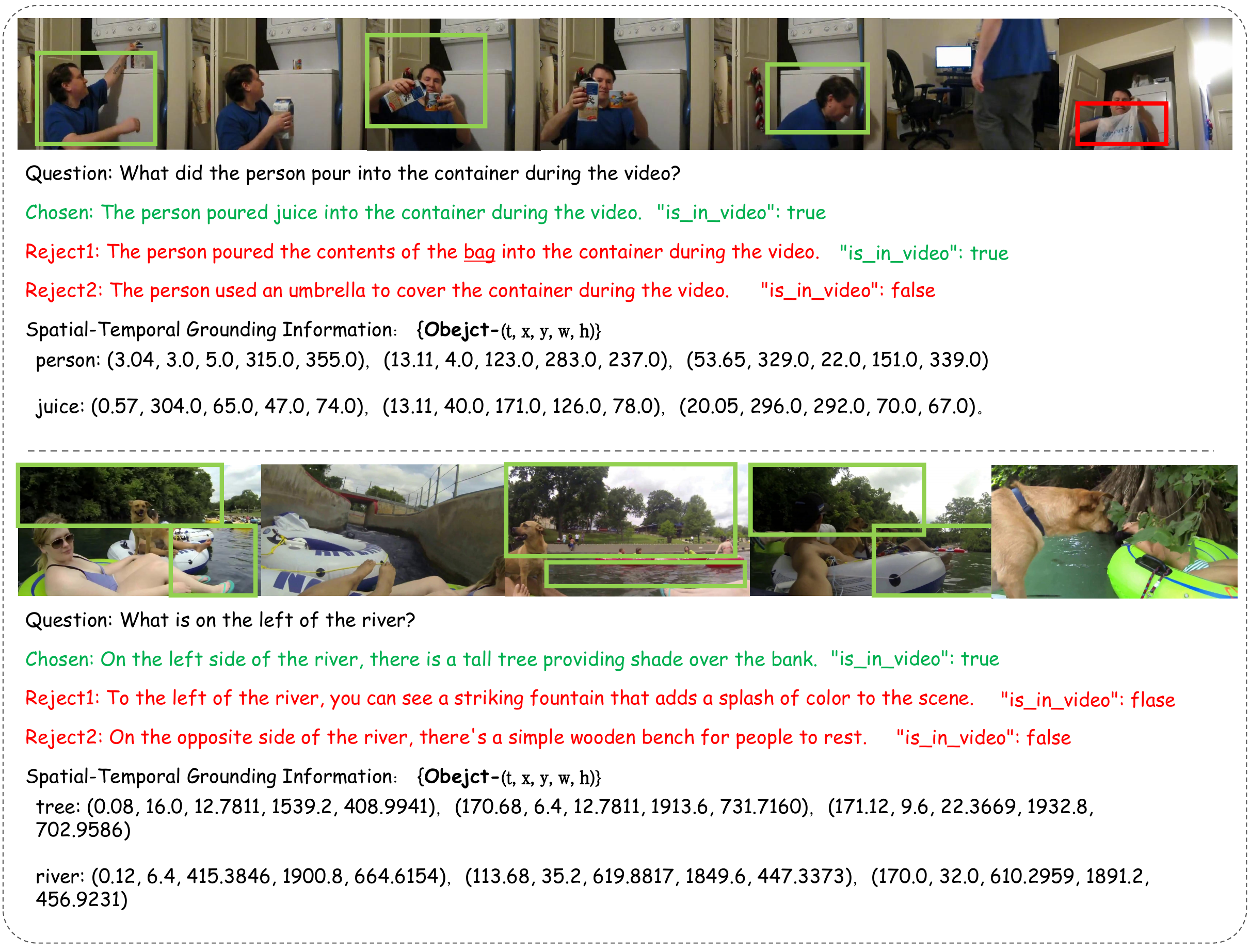}
   \vspace{-1em}
   \caption{Perception data samples of VistaDPO-7K.}
   \vspace{-2mm}
   \label{app_fig:perc_data}
\end{figure}

\textbf{VistaDPO-7K Sample Demonstration.} We show examples of constructed VistaDPO-7K from temporal samples in Figure~\ref{app_fig:temp_data} and perception samples in Figure~\ref{app_fig:perc_data}.

%%%%%%%%%%%%%%%%%%%%%%%%%%%%%%%%%%%%%%%%%%%%%%%%%%%%%%%%%%%%%%%%%%%%%%%%%%%%%%%
%%%%%%%%%%%%%%%%%%%%%%%%%%%%%%%%%%%%%%%%%%%%%%%%%%%%%%%%%%%%%%%%%%%%%%%%%%%%%%%
% APPENDIX
%%%%%%%%%%%%%%%%%%%%%%%%%%%%%%%%%%%%%%%%%%%%%%%%%%%%%%%%%%%%%%%%%%%%%%%%%%%%%%%
%%%%%%%%%%%%%%%%%%%%%%%%%%%%%%%%%%%%%%%%%%%%%%%%%%%%%%%%%%%%%%%%%%%%%%%%%%%%%%%
%%%%%%%%%%%%%%%%%%%%%%%%%%%%%%%%%%%%%%%%%%%%%%%%%%%%%%%%%%%%%%%%%%%%%%%%%%%%%%%
%%%%%%%%%%%%%%%%%%%%%%%%%%%%%%%%%%%%%%%%%%%%%%%%%%%%%%%%%%%%%%%%%%%%%%%%%%%%%%%

\end{document}